\documentclass{article}
\PassOptionsToPackage{numbers, compress}{natbib}

\usepackage[dblblindworkshop, final]{neurips_2026}

\usepackage[utf8]{inputenc} 
\usepackage[T1]{fontenc}    
\usepackage[hidelinks]{hyperref}       
\usepackage{url}            
\usepackage{booktabs}       
\usepackage{graphicx}       
\usepackage{multirow}       
\usepackage{amsmath}        
\usepackage{amsfonts}       
\usepackage{nicefrac}       
\usepackage{microtype}      
\usepackage{xcolor}         
\usepackage{tabularx}

\usepackage[toc,page,header]{appendix}
\usepackage{minitoc}

\doparttoc
\faketableofcontents

\workshoptitle{Representations for the Physical Sciences Workshop}
\title{Robust Transfer Learning for Paper ECG Recognition}

\author{
  Yinghao Xie$^{*}$ \quad
  Zhenbang Dai$^{1,*}$ \quad
  Haojun Wang$^{2,3}$ \quad
  Jinyu Cai \quad
  Fabio Bonassi$^{1}$ \\
  \bfseries
  Hongwu Chen$^{2,3}$ \quad
  Johan Sundstr\"{o}m$^{1}$ \quad
  Jiawei Li$^{1,\dagger}$ \quad
  Ant\^{o}nio H. Ribeiro$^{1}$ \\[0.6em]
  $^{1}$Uppsala University, Sweden \quad
  $^{2}$Nanjing Medical University, China \\
  $^{3}$The First Affiliated Hospital with Nanjing Medical University, China
}

\begin{document}

{\renewcommand{\thefootnote}{\fnsymbol{footnote}}
\footnotetext[1]{Equal contribution.}
\footnotetext[2]{Correspondence: \texttt{jiawei.li@it.uu.se}}
}

\maketitle

\begin{abstract}
Paper ECG recognition is challenging because real-world ECG images vary in layout, physical artifacts, and label availability. We introduce RobECG-CL, a rank-aware contrastive learning framework for robust paper ECG representation learning. Starting from standard 12-lead ECG recordings, we construct progressively degraded paper ECG views with heterogeneous layouts and train the model to balance same-recording invariance with degradation-aware ordering. Across synthetic stress tests on CODE-II and EchoNext, RobECG-CL improves robustness under severe degradation and few-shot transfer, outperforming contrastive learning baselines and surpassing the waveform-based foundation model, ECG-FM, in the 1\% labeled setting. On 312 samples of hospital data with 37 labels, RobECG-CL achieves the best macro AUROC.
\end{abstract}

\section{Introduction and related work}

Paper electrocardiograms (paper ECGs) are ECG recordings that are printed, scanned, or photographed after acquisition. Because ECG machines and hospital archiving systems often retain only reports or images, the original digital waveform may be discarded or overwritten, leaving the paper ECG as the only available record.	

Recognizing paper ECGs is difficult because the model should be robust to both clinical and visual variation. In real-world archives, ECG images vary across lead layouts, rhythm strips, synchronous or asynchronous displays, paper handling artifacts, and annotation availability. These factors make paper ECG recognition a transfer problem: a good representation learned by deep learning models should preserve ECG morphology while remaining stable under layout shift, image degradation, and few-shot downstream supervision.

Large ECG foundation models have shown strong transfer ability from digital waveforms \cite{li2024electrocardiogram,mckeen2024ecg,gu2026cardiac}, and image-based ECG models can predict clinical labels directly from rendered or scanned ECG images \cite{sangha2022automated,reyna2024challenge,reyna2024ecgimage}. The PhysioNet Challenge 2024 focused on ECG image digitization and benchmarked classification as a related task \cite{reyna2024challenge}. However, waveform foundation models cannot be applied when only paper ECG images are available, and the robustness of image-based models under degradation, heterogeneous layouts, and limited labels remains unclear. Our contributions are:

\textbf{(a) We construct a progressive degradation pipeline for the training and evaluation.} We render clean paper ECGs with 3$\times$4 and 6$\times$2 layouts, synchronous or asynchronous displays, and lead-II rhythm strips, then generate matched views with increasing degradation severity.

\textbf{(b) We propose RobECG-CL for robust representation learning.} RobECG-CL uses rank-aware contrastive learning to keep degraded views from the same ECG close in representation space, while encouraging mildly degraded views to stay closer to the clean anchor than severely degraded views.

\textbf{(c) We evaluate transfer under synthetic stress tests.} Across sex prediction and structural heart disease (SHD) classification, RobECG-CL achieves the strongest overall robustness across four degradation severity levels and outperforms waveform foundation models in the 1\% few-shot setting. 

\textbf{(d) We evaluate transfer under real-world stress tests.} We evaluated RobECG-CL on a small paper ECG dataset from a hospital with 37  labels. RobECG-CL achieves the best macro AUROC on all labels and the second best on the top 12 most frequent labels.

\section{Study design and methods}

\begin{figure}[t]
    \centering
    \includegraphics[width=\textwidth]{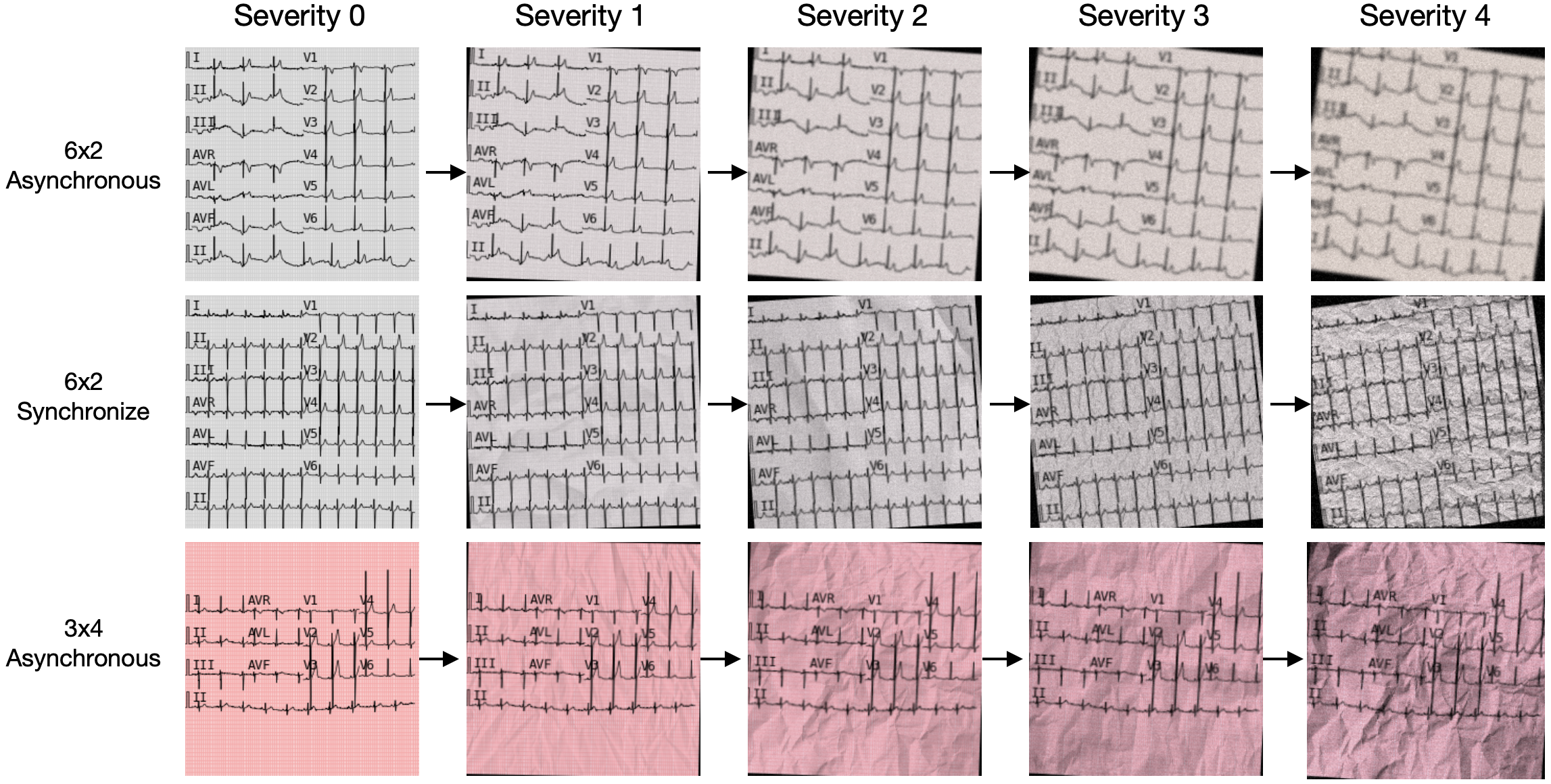}
    \caption{Progressive degradation. For each ECG recording, we first render a clean paper ECG using a randomly sampled layout, and then generate progressively degraded views by applying cumulative visual corruptions, including rotation, geometric perturbation, blur, color shift, noise, and artifacts.}
    \label{fig:progressive-degradation}
\end{figure}

\textbf{Study design.}
We study whether self-supervised pretraining on progressively degraded paper ECGs improves transfer under layout shift, image degradation, and limited labels. For each standard 12-lead ECG recording, we first render a clean paper ECG image $\mathbf{x}_i^{(0)}\in\mathbb{R}^{3\times 224\times 224}$ using a rendering configuration $\ell_i$. The $\ell_i$ is sampled once and shared by all views from the same recording: we sample a 3$\times$4 or 6$\times$2 layout with equal probability, include a lead-II rhythm strip at the bottom, and use synchronous or asynchronous lead display with equal probability. For upstream pretraining, we use PTB-XL \cite{wagner2020ptb}. For controlled transfer, we apply the same rendering and degradation definitions to CODE-II for sex classification and age regression \cite{abreu2025code}, and to EchoNext for SHD classification \cite{PhysioNet-echonext-1.1.0}. 

\textbf{Progressive degradation.}
As illustrated in Figure~\ref{fig:progressive-degradation}, we generate an ordered set of degradation severity levels from each clean-rendered view:
\begin{equation}
\begin{aligned}
    \mathbf{u}_i^{(k)}
    &=
    (1-\mathbf{M}_i^{(k)})\odot
    \mathcal{C}_i^{(k)}
    \!\left(\mathcal{W}_{\mathbf{A}_i^{(k)}}(\mathbf{x}_i^{(0)})\right)
    +\mathbf{M}_i^{(k)}\odot\mathbf{P}_i^{(k)},\\
    \mathbf{x}_i^{(k)}
    &=
    \operatorname{clip}\!\left[
    \mathbf{K}_i^{(k)}\ast\mathbf{u}_i^{(k)}
    +\boldsymbol{\epsilon}_i^{(k)}
    \right],
    \qquad r(\mathbf{x}_i^{(k)})=k .
\end{aligned}
    \label{eq:progressive-degradation}
\end{equation}
Here,  $\mathcal{W}_{\mathbf{A}_i^{(k)}}$ is a geometric warp parameterized by an affine matrix $\mathbf{A}_i^{(k)}$, covering rotation and cropping. $\mathcal{C}_i^{(k)}$ is a sampled color transform, $\mathbf{M}_i^{(k)}$ is an artifact mask, $\mathbf{P}_i^{(k)}$ is a sampled paper artifact texture, $\mathbf{K}_i^{(k)}$ is a blur kernel, $\ast$ denotes convolution, $\odot$ denotes element-wise multiplication, and $\boldsymbol{\epsilon}_i^{(k)}$ is additive acquisition noise. 
The degradation state is updated cumulatively. Specifically,
\begin{equation}
\begin{aligned}
    \mathbf{A}_i^{(k)}
    &=\Delta\mathbf{A}_i^{(k)}\mathbf{A}_i^{(k-1)},&
    \mathcal{C}_i^{(k)}
    &=\Delta\mathcal{C}_i^{(k)}\circ\mathcal{C}_i^{(k-1)},\\
    \mathbf{M}_i^{(k)}
    &=\mathbf{M}_i^{(k-1)}+\Delta\mathbf{M}_i^{(k)},&
    \beta_i^{(k)}
    &=\beta_i^{(k-1)}+\Delta\beta_i^{(k)},\\
    \sigma_i^{(k)}
    &=\sigma_i^{(k-1)}+\Delta\sigma_i^{(k)},&
    \mathbf{K}_i^{(k)}
    &=\operatorname{BlurKernel}(\beta_i^{(k)}).
\end{aligned}
    \label{eq:degradation-state-update}
\end{equation}
Here, $\Delta\mathbf{A}_i^{(k)}$ and $\Delta\mathcal{C}_i^{(k)}$ are newly sampled geometric and color increments, while $\Delta\mathbf{M}_i^{(k)}$, $\Delta\beta_i^{(k)}$, and $\Delta\sigma_i^{(k)}$ increase the artifact, blur, and noise states. Thus, larger $k$ denotes stronger degradation.

\textbf{Rank-aware contrastive pretraining.}
RobECG-CL uses a CvT-13 image encoder $f_\theta$ and a two-layer MLP $g_\phi$ to map each view to a normalized embedding $\mathbf{z}_i^{(k)}=g_\phi(f_\theta(\mathbf{x}_i^{(k)}))$. For each recording, the clean-rendered view is used as the anchor, all degraded views from the same recording are treated as positives, and views from other recordings are negatives. Let $s(\cdot, \cdot)$ denote cosine similarity and let $\mathcal{V}_i$ be the degraded views of recording $i$. The set-level multi-positive InfoNCE loss is
\begin{equation}
    \mathcal{L}_{\mathrm{set}}
    =
    -\frac{1}{B}\sum_{i=1}^{B}
    \log
    \frac{
    \sum_{\mathbf{v}\in\mathcal{V}_i}
    \exp(s(\mathbf{z}_i^{(0)},\mathbf{v})/\tau)
    }{
    \sum_{\mathbf{q}\in\mathcal{Q}_i}
    \exp(s(\mathbf{z}_i^{(0)},\mathbf{q})/\tau)
    },
    \label{eq:set-infonce}
\end{equation}
where $\mathcal{Q}_i$ contains all candidate views in the mini-batch $B$ except the anchor. 
The InfoNCE loss treats all degraded views from the same ECG as equally positive, although severe degradation can obscure fine ECG morphology which might be crucial for diagnosis. We therefore encourage a degradation-aware ordinal representation, where mildly degraded views stay closer to the clean-rendered anchor than severely degraded views.
 Inspired by rank-aware learning \cite{zha2023rank}, we construct ordinal pairs using two distances: a Canny-based geometric distance $d_g$ and a CIELAB color-based distance $d_c$. A pair $(p,q)$ is used as an ordinal edge if view $p$ is consistently closer to the anchor than view $q$:
\begin{equation}
    \mathcal{E}_i=
    \left\{
    (p,q):
    d_g(\mathbf{x}_i^{(0)},\mathbf{x}_i^{(p)})
    \le(1-\eta)d_g(\mathbf{x}_i^{(0)},\mathbf{x}_i^{(q)}),
    \;
    d_c(\mathbf{x}_i^{(0)},\mathbf{x}_i^{(p)})
    \le(1-\eta)d_c(\mathbf{x}_i^{(0)},\mathbf{x}_i^{(q)})
    \right\}.
    \label{eq:ordinal-edges}
\end{equation}
The less degraded view should be more similar to the anchor than the more degraded view:
\begin{equation}
    \mathcal{L}_{\mathrm{rank}}
    =
    \frac{1}{|\mathcal{I}|}
    \sum_{i\in\mathcal{I}}
    \frac{1}{|\mathcal{E}_i|}
    \sum_{(p,q)\in\mathcal{E}_i}
    \left[
    m-s(\mathbf{z}_i^{(0)},\mathbf{z}_i^{(p)})
    +s(\mathbf{z}_i^{(0)},\mathbf{z}_i^{(q)})
    \right]_{+}^{2},
    \label{eq:rank-loss}
\end{equation}
where $\mathcal{I}=\{i:|\mathcal{E}_i|>0\}$, $[\cdot]_+=\max(\cdot,0)$, and $m$ is a hyperparameter. The final objective is
$\mathcal{L}=\mathcal{L}_{\mathrm{set}}+\lambda\mathcal{L}_{\mathrm{rank}} .
$ The model learns both invariance across degradation severity levels and a weak ordering of image degradation.

\begin{table*}[t]
	\centering
	\scriptsize
	\setlength{\tabcolsep}{2.4pt}
	\caption{AUROC across severity levels on CODE-II sex classification. Higher severity levels denote progressively stronger degradation. Values are mean $\pm$ standard deviation over five runs.}
	\label{tab:codeii-sex}
	\resizebox{\textwidth}{!}{%
		\begin{tabular}{lcccccccccccc}
			\toprule
			\multirow{2}{*}{Method} & \multicolumn{2}{c}{Clean Paper ECG} & \multicolumn{2}{c}{Severity 1} & \multicolumn{2}{c}{Severity 2} & \multicolumn{2}{c}{Severity 3} & \multicolumn{2}{c}{Severity 4} & \multicolumn{2}{c}{Avg.} \\
			\cmidrule(lr){2-3}\cmidrule(lr){4-5}\cmidrule(lr){6-7}\cmidrule(lr){8-9}\cmidrule(lr){10-11}\cmidrule(lr){12-13}
			& 1\% & 10\% & 1\% & 10\% & 1\% & 10\% & 1\% & 10\% & 1\% & 10\% & 1\% & 10\% \\
			\midrule
			\multicolumn{13}{l}{\emph{PTB-XL $\rightarrow$ downstream}} \\
			RobECG-CL & \textbf{0.766 $\pm$ 0.006} & 0.798 $\pm$ 0.003 & \textbf{0.766 $\pm$ 0.005} & 0.797 $\pm$ 0.003 & \textbf{0.763 $\pm$ 0.005} & \textbf{0.793 $\pm$ 0.003} & \textbf{0.758 $\pm$ 0.006} & \textbf{0.786 $\pm$ 0.003} & \textbf{0.742 $\pm$ 0.005} & \textbf{0.768 $\pm$ 0.003} & \textbf{0.759 $\pm$ 0.005} & \textbf{0.788 $\pm$ 0.003} \\
			MoCo & 0.756 $\pm$ 0.007 & \textbf{0.806 $\pm$ 0.002} & 0.740 $\pm$ 0.006 & \textbf{0.798 $\pm$ 0.002} & 0.717 $\pm$ 0.009 & 0.776 $\pm$ 0.004 & 0.682 $\pm$ 0.013 & 0.745 $\pm$ 0.004 & 0.639 $\pm$ 0.016 & 0.705 $\pm$ 0.006 & 0.710 $\pm$ 0.007 & 0.768 $\pm$ 0.003 \\
			DINO & 0.707 $\pm$ 0.013 & 0.782 $\pm$ 0.003 & 0.706 $\pm$ 0.014 & 0.780 $\pm$ 0.003 & 0.684 $\pm$ 0.019 & 0.766 $\pm$ 0.003 & 0.651 $\pm$ 0.019 & 0.741 $\pm$ 0.004 & 0.614 $\pm$ 0.017 & 0.704 $\pm$ 0.005 & 0.675 $\pm$ 0.015 & 0.756 $\pm$ 0.003 \\
			BYOL & 0.739 $\pm$ 0.006 & 0.803 $\pm$ 0.001 & 0.724 $\pm$ 0.007 & 0.793 $\pm$ 0.001 & 0.698 $\pm$ 0.009 & 0.772 $\pm$ 0.003 & 0.663 $\pm$ 0.011 & 0.743 $\pm$ 0.003 & 0.623 $\pm$ 0.013 & 0.708 $\pm$ 0.003 & 0.693 $\pm$ 0.008 & 0.766 $\pm$ 0.002 \\
			SimCLR & 0.723 $\pm$ 0.010 & 0.754 $\pm$ 0.003 & 0.716 $\pm$ 0.008 & 0.750 $\pm$ 0.003 & 0.698 $\pm$ 0.007 & 0.741 $\pm$ 0.004 & 0.669 $\pm$ 0.006 & 0.723 $\pm$ 0.004 & 0.633 $\pm$ 0.007 & 0.689 $\pm$ 0.005 & 0.690 $\pm$ 0.006 & 0.732 $\pm$ 0.003 \\
			\multicolumn{13}{l}{\emph{Pretrained (image-based) $\rightarrow$ downstream}} \\
			PULSE & 0.770 $\pm$ 0.011 & 0.822 $\pm$  0.003& 0.764 $\pm$ 0.013 & 0.816 $\pm$  0.002& \textbf{0.750 $\pm$ 0.016}& 0.802 $\pm$  0.004& 0.727 $\pm$ 0.017 & 0.781 $\pm$ 0.005& 0.695 $\pm$ 0.017 & 0.752 $\pm$  0.004& 0.742 $\pm$ 0.014 & 0.795 $\pm$  0.003\\
			GEM & \textbf{0.772 $\pm$ 0.009} & \textbf{0.821 $\pm$ 0.009}& \textbf{0.766 $\pm$ 0.010} & \textbf{0.818 $\pm$  0.007}& 0.750 $\pm$ 0.013& \textbf{0.805 $\pm$  0.006}& \textbf{0.729 $\pm$ 0.013} & \textbf{0.786 $\pm$ 0.006}& \textbf{0.697 $\pm$ 0.013} & \textbf{0.757 $\pm$  0.007}& \textbf{0.743 $\pm$ 0.010} & \textbf{0.797 $\pm$ 0.007}\\
			\multicolumn{13}{l}{\emph{Pretrained (waveform-based) $\rightarrow$ downstream}} \\
			CSFM & N/A & N/A & N/A & N/A & N/A & N/A & N/A & N/A & N/A & N/A & \textbf{0.743 $\pm$ 0.026} & 0.864 $\pm$ 0.004 \\
			ECGFounder & N/A & N/A & N/A & N/A & N/A & N/A & N/A & N/A & N/A & N/A & 0.644 $\pm$ 0.040 & 0.860 $\pm$ 0.007 \\
			ECG\_FM & N/A & N/A & N/A & N/A & N/A & N/A & N/A & N/A & N/A & N/A & 0.681 $\pm$ 0.034 & \textbf{0.866 $\pm$ 0.001} \\
			\bottomrule
		\end{tabular}%
	}
\end{table*}

\begin{table*}[t]
	\centering
	\scriptsize
	\setlength{\tabcolsep}{2.4pt}
	\caption{AUROC across severity levels on EchoNext SHD classification. Higher severity levels denote progressively stronger degradation. Values are mean $\pm$ standard deviation over five runs.}
	\label{tab:echonext}
	\resizebox{\textwidth}{!}{%
		\begin{tabular}{lcccccccccccc}
			\toprule
			\multirow{2}{*}{Method} & \multicolumn{2}{c}{Clean Paper ECG} & \multicolumn{2}{c}{Severity 1} & \multicolumn{2}{c}{Severity 2} & \multicolumn{2}{c}{Severity 3} & \multicolumn{2}{c}{Severity 4} & \multicolumn{2}{c}{Avg.} \\
			\cmidrule(lr){2-3}\cmidrule(lr){4-5}\cmidrule(lr){6-7}\cmidrule(lr){8-9}\cmidrule(lr){10-11}\cmidrule(lr){12-13}
			& 1\% & 10\% & 1\% & 10\% & 1\% & 10\% & 1\% & 10\% & 1\% & 10\% & 1\% & 10\% \\
			\midrule
			\multicolumn{13}{l}{\emph{PTB-XL $\rightarrow$ downstream}} \\
			RobECG-CL & 0.692 $\pm$ 0.006 & 0.738 $\pm$ 0.002 & 0.691 $\pm$ 0.006 & 0.735 $\pm$ 0.002 & \textbf{0.689 $\pm$ 0.006} & 0.734 $\pm$ 0.002 & \textbf{0.684 $\pm$ 0.006} & \textbf{0.729 $\pm$ 0.002} & \textbf{0.673 $\pm$ 0.006} & \textbf{0.715 $\pm$ 0.002} & \textbf{0.686 $\pm$ 0.006} & \textbf{0.730 $\pm$ 0.002} \\
			MoCo & \textbf{0.722 $\pm$ 0.002} & \textbf{0.768 $\pm$ 0.004} & \textbf{0.709 $\pm$ 0.004} & 0.756 $\pm$ 0.005 & 0.681 $\pm$ 0.007 & 0.726 $\pm$ 0.007 & 0.642 $\pm$ 0.010 & 0.690 $\pm$ 0.006 & 0.600 $\pm$ 0.010 & 0.644 $\pm$ 0.007 & 0.671 $\pm$ 0.005 & 0.717 $\pm$ 0.005 \\
			DINO & 0.709 $\pm$ 0.005 & 0.766 $\pm$ 0.004 & 0.695 $\pm$ 0.006 & \textbf{0.758 $\pm$ 0.004} & 0.670 $\pm$ 0.011 & \textbf{0.743 $\pm$ 0.005} & 0.631 $\pm$ 0.012 & 0.708 $\pm$ 0.005 & 0.594 $\pm$ 0.014 & 0.667 $\pm$ 0.004 & 0.660 $\pm$ 0.009 & 0.728 $\pm$ 0.004 \\
			BYOL & 0.711 $\pm$ 0.006 & 0.764 $\pm$ 0.004 & 0.689 $\pm$ 0.007 & 0.749 $\pm$ 0.004 & 0.663 $\pm$ 0.008 & 0.723 $\pm$ 0.004 & 0.626 $\pm$ 0.010 & 0.694 $\pm$ 0.005 & 0.586 $\pm$ 0.012 & 0.657 $\pm$ 0.006 & 0.655 $\pm$ 0.006 & 0.717 $\pm$ 0.004 \\
			SimCLR & 0.664 $\pm$ 0.010 & 0.698 $\pm$ 0.003 & 0.657 $\pm$ 0.010 & 0.691 $\pm$ 0.003 & 0.643 $\pm$ 0.012 & 0.676 $\pm$ 0.003 & 0.620 $\pm$ 0.010 & 0.651 $\pm$ 0.005 & 0.582 $\pm$ 0.008 & 0.615 $\pm$ 0.003 & 0.633 $\pm$ 0.010 & 0.666 $\pm$ 0.003 \\
			\multicolumn{13}{l}{\emph{Pretrained (image-based) $\rightarrow$ downstream}} \\
			PULSE & \textbf{0.766 $\pm$ 0.010} & \textbf{0.793 $\pm$  0.008}& \textbf{0.751 $\pm$ 0.010} & 0.785 $\pm$  0.008& \textbf{0.734 $\pm$ 0.009} & 0.774 $\pm$ 0.008& \textbf{0.712 $\pm$ 0.012} & 0.758 $\pm$  0.007& \textbf{0.677 $\pm$ 0.010} & \textbf{0.726 $\pm$ 0.007}& \textbf{0.728 $\pm$ 0.009} & \textbf{0.767 $\pm$ 0.006}\\
			GEM & 0.759 $\pm$ 0.014 & 0.792 $\pm$  0.009& 0.745 $\pm$ 0.013 & \textbf{0.786 $\pm$  0.009}& 0.727 $\pm$ 0.014 & \textbf{0.775 $\pm$  0.010}& 0.705 $\pm$ 0.016 & \textbf{0.759 $\pm$  0.009}& 0.667 $\pm$ 0.018 & 0.721 $\pm$ 0.011& 0.721 $\pm$ 0.013 & 0.766 $\pm$ 0.009\\
			\multicolumn{13}{l}{\emph{Pretrained (waveform-based) $\rightarrow$ downstream}} \\
			CSFM & N/A & N/A & N/A & N/A & N/A & N/A & N/A & N/A & N/A & N/A & 0.651 $\pm$ 0.078 & 0.768 $\pm$ 0.003 \\
			ECGFounder & N/A & N/A & N/A & N/A & N/A & N/A & N/A & N/A & N/A & N/A & 0.615 $\pm$ 0.056 & 0.764 $\pm$ 0.004 \\
			ECG\_FM & N/A & N/A & N/A & N/A & N/A & N/A & N/A & N/A & N/A & N/A & \textbf{0.657 $\pm$ 0.095} & \textbf{0.795 $\pm$ 0.004} \\
			\bottomrule
		\end{tabular}%
	}
\end{table*}

\begin{table*}[h]
	\centering
	\scriptsize
	\setlength{\tabcolsep}{1.8pt}
	\caption{Per-label AUROC and F1 on the 12 diagnostic labels with more than 20 samples in a real-world paper ECG dataset. Results are pooled out-of-fold performance under 5-fold evaluation. Full results on 37 labels are provided in Supplementary Table~\ref{tab:real-world-hospital-all37}.}
	\label{tab:real-world-hospital}
	\resizebox{\textwidth}{!}{%
		\begin{tabular}{lcccccccccccccccccccccccccc}
			\toprule
			\multirow{2}{*}{Method} & \multicolumn{2}{c}{1AVB} & \multicolumn{2}{c}{AF} & \multicolumn{2}{c}{CRBBB} & \multicolumn{2}{c}{IRBBB} & \multicolumn{2}{c}{ISC} & \multicolumn{2}{c}{LAE} & \multicolumn{2}{c}{LVH} & \multicolumn{2}{c}{NORM} & \multicolumn{2}{c}{OMI} & \multicolumn{2}{c}{PAC} & \multicolumn{2}{c}{PVC} & \multicolumn{2}{c}{SB} & \multicolumn{2}{c}{Avg.} \\
			\cmidrule(lr){2-3}\cmidrule(lr){4-5}\cmidrule(lr){6-7}\cmidrule(lr){8-9}\cmidrule(lr){10-11}\cmidrule(lr){12-13}\cmidrule(lr){14-15}\cmidrule(lr){16-17}\cmidrule(lr){18-19}\cmidrule(lr){20-21}\cmidrule(lr){22-23}\cmidrule(lr){24-25}\cmidrule(lr){26-27}
			& AUC & F1 & AUC & F1 & AUC & F1 & AUC & F1 & AUC & F1 & AUC & F1 & AUC & F1 & AUC & F1 & AUC & F1 & AUC & F1 & AUC & F1 & AUC & F1 & AUC & F1 \\
			\midrule
			RobECG-CL & \textbf{0.581} & \textbf{0.212} & 0.753 & 0.278 & \textbf{0.928} & \textbf{0.691} & \underline{0.626} & 0.140 & 0.653 & 0.271 & \textbf{0.830} & \textbf{0.429} & \underline{0.952} & \underline{0.615} & \textbf{0.841} & \textbf{0.299} & 0.557 & 0.177 & 0.699 & \textbf{0.219} & 0.842 & 0.590 & \textbf{0.912} & \textbf{0.635} & \underline{0.765} & \textbf{0.380} \\
			MoCo & \underline{0.541} & \underline{0.171} & 0.752 & \textbf{0.393} & 0.877 & 0.597 & \textbf{0.645} & 0.192 & 0.736 & \underline{0.346} & 0.777 & 0.226 & 0.944 & 0.516 & 0.755 & 0.246 & \underline{0.628} & \underline{0.186} & \underline{0.764} & 0.156 & 0.899 & 0.548 & \underline{0.892} & \underline{0.558} & \textbf{0.767} & \underline{0.345} \\
			DINO & 0.499 & 0.171 & \underline{0.778} & \underline{0.295} & 0.880 & 0.571 & 0.604 & \textbf{0.213} & \underline{0.750} & 0.293 & 0.702 & 0.222 & 0.877 & 0.324 & \underline{0.839} & \underline{0.293} & \textbf{0.628} & \textbf{0.212} & \textbf{0.799} & 0.203 & 0.903 & 0.635 & 0.848 & 0.467 & 0.759 & 0.325 \\
			BYOL & 0.500 & 0.134 & 0.617 & 0.273 & 0.848 & 0.522 & 0.574 & 0.139 & 0.657 & 0.259 & 0.745 & 0.203 & \textbf{0.960} & \textbf{0.636} & 0.792 & 0.250 & 0.575 & 0.115 & 0.729 & 0.216 & 0.884 & 0.577 & 0.889 & 0.454 & 0.731 & 0.315 \\
			SimCLR & 0.510 & 0.148 & 0.618 & 0.256 & 0.722 & 0.278 & 0.540 & 0.136 & \textbf{0.788} & \textbf{0.351} & \underline{0.790} & 0.189 & 0.928 & 0.415 & 0.778 & 0.233 & 0.598 & 0.182 & 0.691 & \underline{0.216} & 0.813 & 0.463 & 0.840 & 0.500 & 0.718 & 0.281 \\
			PULSE & 0.531 & 0.131 & 0.750 & 0.275 & 0.895 & 0.597 & 0.569 & 0.162 & 0.688 & 0.205 & 0.737 & 0.138 & 0.859 & 0.579 & 0.774 & 0.222 & 0.493 & 0.057 & 0.635 & 0.000 & \textbf{0.915} & \textbf{0.685} & 0.816 & 0.314 & 0.722 & 0.280 \\
			GEM & 0.532 & 0.140 & \textbf{0.818} & 0.244 & \underline{0.908} & \underline{0.677} & 0.607 & \underline{0.194} & 0.726 & 0.270 & 0.721 & \underline{0.231} & 0.816 & 0.474 & 0.791 & 0.118 & 0.514 & 0.179 & 0.635 & 0.138 & \underline{0.908} & \underline{0.683} & 0.811 & 0.406 & 0.732 & 0.313 \\
			\bottomrule
		\end{tabular}%
	}
\end{table*}

\section{Results}
We evaluate against three baseline categories: (1) contrastive learning methods pretrained on PTB-XL (MoCo~\cite{he2020moco}, DINO~\cite{caron2021dino}, BYOL~\cite{grill2020byol}, SimCLR~\cite{chen2020simclr}); (2) pretrained paper ECG foundation models (PULSE~\cite{liu2026teaching}, GEM~\cite{lan2026gem}); and (3) pretrained waveform foundation models (CSFM~\cite{gu2026cardiac}, ECGfounder~\cite{li2024electrocardiogram}, ECG-FM~\cite{mckeen2024ecg}). Transfer learning is assessed on sex and SHD classification using 1\% and 10\% of the training data (Tables~\ref{tab:codeii-sex} and \ref{tab:echonext}), with CODE-II age prediction results deferred to Appendix~\ref{sec:age}. Finally, we validate on 312 real-world paper ECGs from The First Affiliated Hospital with Nanjing Medical University.

\textbf{Robustness under limited labeled data and progressive degradation.} RobECG-CL achieves the best macro AUROC on both sex classification and SHD classification among models pretrained on PTB-XL paper ECGs. The advantage is more evident at severe degradation levels (i.e., severity 4), indicating that rank-aware contrastive learning improves robustness.

\textbf{Comparison with waveform foundation models.} Although paper ECG images provide only an indirect visual observation of the underlying waveform, RobECG-CL outperforms ECG-FM under 1\% training data on both sex and SHD classifications.

\textbf{Comparison with image-based foundation models.} PULSE and GEM show strong transferability to degraded paper ECG images, even at severe degradation levels. However, they underperform RobECG-CL on sex classification and age prediction when using 1\% labeled training data (Table~\ref{tab:codeii-age}).

\textbf{Analysis on Real-World Paper ECGs from a Hospital.}
The images were annotated by an experienced cardiologist and captured from printed ECG records using a RICOH 8001 scanner. Additional details are provided in the Appendix~\ref{sec:appendix_real_world_data}.
As shown in Table~\ref{tab:real-world-hospital}, among the 12 labels with more than 20 positive samples, RobECG-CL achieves the second-best macro AUROC and the best macro F1 compared with other contrastive learning baselines and image-based foundation models.

\begin{figure}[h]
	\centering
	\includegraphics[width=\textwidth]{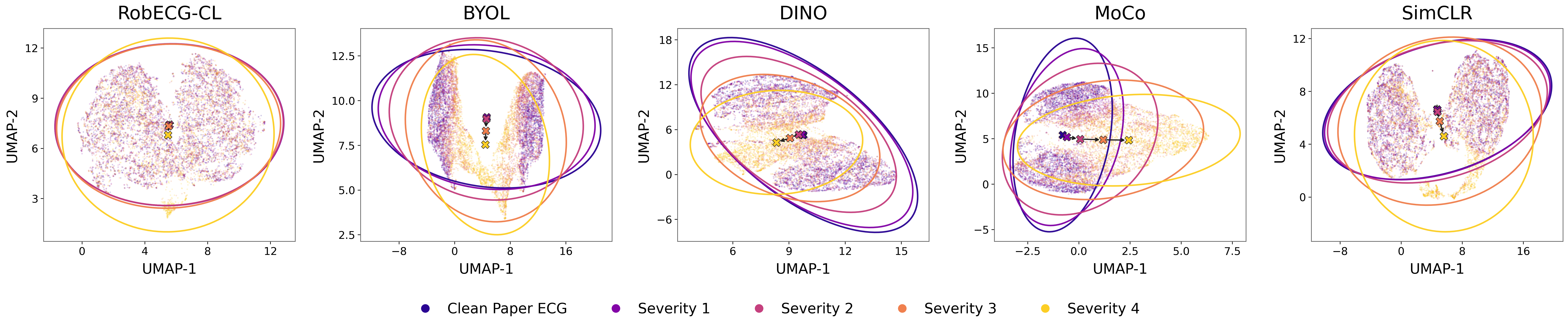}
	\caption{UMAP visualization of representations for sex prediction with 10\% labeled training data. RobECG-CL learns degradation-robust representations while preserving the relative structure among degraded views. The severely corrupted views are not forced to be close to the clean view.}
	\label{fig:representation-drift}
\end{figure}

\textbf{Visualization of Robust Representations.} As illustrated in Figure~\ref{fig:representation-drift}, our method generally aligns the latent representations closely with the clean anchor across varying degradation severities. It also flexibly permits a small subset of severely degraded samples to deviate to avoid unnatural alignments.

\section{Conclusion and limitation}

In this study, we propose RobECG-CL, a robust self-supervised learning framework designed to extract degradation-aware representations from paper ECG images. 
By leveraging a synthetic dataset with progressive degradations, our method effectively aligns features across varying severity levels without forcing unnatural alignments for severely degraded samples. Consequently, RobECG-CL demonstrates superior classification performance and strong transferability to real-world clinical records. 
A limitation of our study is that the baselines pretrained on PTB-XL rely on augmentation pipelines for clean ECGs, rather than treating progressively degraded views as positive pairs. Moreover, some degraded views might lose the clinical features required for accurate diagnosis, and we did not manually filter these cases. Finally, our real-world validation set is limited to a single center.

\section{Acknowledgments and Disclosure of Funding}
\label{sec:aknowledgment}
Jiawei Li and Antônio H. Ribeiro are financially supported by the eSSENCE and SciLifeLab, with the
project "Digital Biomarkers from the Electrocardiogram using Artificial Intelligence";
and, by the Wallenberg AI, Autonomous Systems and Software Program (WASP) funded by Knut and Alice Wallenberg Foundation. 
This project has received funding from the European Research Council (ERC) under the European Union's Horizon Europe research and innovation programme through grant agreement no. 101054643.
Computations were enabled by resources provided by the National Academic Infrastructure for Supercomputing in Sweden (NAISS), partially funded by the Swedish Research Council through grant agreement no. 2022-06725.

\bibliography{Neurips_workshop}
\bibliographystyle{conference}
	
\newpage
\appendix

\part{}

\parttoc

\newpage

\section{Supplementary results}
\subsection{Supplementary age prediction results}
\label{sec:age}
RobECG-CL achieves the best age prediction performance among the PTB-XL-pretrained methods under both the 1\% and 10\% training settings (Table~\ref{tab:codeii-age}).  Increasing the labeled training data from 1\% to 10\% reduces its average MAE and increases the average $R^2$. Additionally, RobECG-CL exhibits a smaller increase in MAE as degradation severity increases. Compared with PULSE and GEM, RobECG-CL achieves better performance in the more challenging 1\% labeled training data setting and remains competitive when 10\% labeled training data are available, although PULSE and GEM obtain lower absolute errors in the latter setting. 

\begin{table*}[h]
	\centering
	\scriptsize
	\setlength{\tabcolsep}{1.6pt}
	\caption{Transfer performance across degradation severity levels on CODE-II age prediction. Higher severity levels denote progressively stronger degradation. Values are mean $\pm$ standard deviation over five runs.}
	\label{tab:codeii-age}
	\resizebox{\textwidth}{!}{%
		\begin{tabular}{lcccccccccccccccc}
			\toprule
			\multirow{2}{*}{Method} & \multicolumn{2}{c}{Clean Paper ECG MAE} & \multicolumn{2}{c}{Severity 1 MAE} & \multicolumn{2}{c}{Severity 2 MAE} & \multicolumn{2}{c}{Severity 3 MAE} & \multicolumn{2}{c}{Severity 4 MAE} & \multicolumn{2}{c}{Avg. MAE} & \multicolumn{2}{c}{Avg. RMSE} & \multicolumn{2}{c}{Avg. $R^2$} \\
			\cmidrule(lr){2-3}\cmidrule(lr){4-5}\cmidrule(lr){6-7}\cmidrule(lr){8-9}\cmidrule(lr){10-11}\cmidrule(lr){12-13}\cmidrule(lr){14-15}\cmidrule(lr){16-17}
			& 1\% & 10\% & 1\% & 10\% & 1\% & 10\% & 1\% & 10\% & 1\% & 10\% & 1\% & 10\% & 1\% & 10\% & 1\% & 10\% \\
			\midrule
			RobECG-CL & 12.66 $\pm$ 0.22& 11.75 $\pm$ 0.15 & \textbf{12.67 $\pm$ 0.19} & 11.78 $\pm$ 0.17 & \textbf{12.76 $\pm$ 0.21} & 11.87 $\pm$ 0.18 & \textbf{12.90 $\pm$ 0.23} & 12.07 $\pm$ 0.18 & \textbf{13.16 $\pm$ 0.25} & 12.45 $\pm$ 0.19 & \textbf{12.83 $\pm$ 0.21} & 11.98 $\pm$ 0.17 & \textbf{16.00 $\pm$ 0.31} & 15.07 $\pm$ 0.23 & \textbf{0.102 $\pm$ 0.035} & 0.204 $\pm$ 0.024 \\
			MoCo & 13.10 $\pm$ 0.10 & 12.18 $\pm$ 0.08 & 13.00 $\pm$ 0.09 & 12.26 $\pm$ 0.07 & 13.25 $\pm$ 0.13 & 12.50 $\pm$ 0.05 & 13.61 $\pm$ 0.22 & 12.87 $\pm$ 0.06 & 13.88 $\pm$ 0.22 & 13.22 $\pm$ 0.06 & 13.37 $\pm$ 0.14 & 12.60 $\pm$ 0.05 & 16.48 $\pm$ 0.16 & 15.66 $\pm$ 0.07 & 0.047 $\pm$ 0.019 & 0.140 $\pm$ 0.007 \\
			DINO & 13.28 $\pm$ 0.20 & 12.10 $\pm$ 0.05 & 13.27 $\pm$ 0.22 & 12.14 $\pm$ 0.05 & 13.42 $\pm$ 0.23 & 12.38 $\pm$ 0.05 & 13.66 $\pm$ 0.15 & 12.76 $\pm$ 0.06 & 13.90 $\pm$ 0.10 & 13.17 $\pm$ 0.07 & 13.51 $\pm$ 0.17 & 12.51 $\pm$ 0.05 & 16.64 $\pm$ 0.18 & 15.47 $\pm$ 0.06 & 0.028 $\pm$ 0.021 & 0.160 $\pm$ 0.006 \\
			BYOL & 13.14 $\pm$ 0.12 & 12.12 $\pm$ 0.05 & 13.25 $\pm$ 0.15 & 12.26 $\pm$ 0.04 & 13.46 $\pm$ 0.20 & 12.50 $\pm$ 0.03 & 13.69 $\pm$ 0.19 & 12.87 $\pm$ 0.02 & 13.88 $\pm$ 0.13 & 13.22 $\pm$ 0.02 & 13.48 $\pm$ 0.15 & 12.59 $\pm$ 0.01 & 16.66 $\pm$ 0.16 & 15.65 $\pm$ 0.03 & 0.027 $\pm$ 0.019 & 0.141 $\pm$ 0.003 \\
			SimCLR & 13.24 $\pm$ 0.09 & 12.60 $\pm$ 0.08 & 13.28 $\pm$ 0.06 & 12.65 $\pm$ 0.09 & 13.52 $\pm$ 0.07 & 12.75 $\pm$ 0.07 & 13.72 $\pm$ 0.09 & 13.02 $\pm$ 0.03 & 13.86 $\pm$ 0.13 & 13.31 $\pm$ 0.06 & 13.52 $\pm$ 0.08 & 12.87 $\pm$ 0.05 & 16.68 $\pm$ 0.10 & 15.94 $\pm$ 0.06 & 0.025 $\pm$ 0.011 & 0.108 $\pm$ 0.007 \\
			PULSE & \textbf{12.65 $\pm$ 0.39} & 10.98 $\pm$  0.19& 12.74 $\pm$ 0.32 & 11.13 $\pm$  0.22& 12.89 $\pm$ 0.27 &\textbf{11.38 $\pm$ 0.19}& 13.18 $\pm$ 0.22 & \textbf{11.72 $\pm$ 0.17}& 13.65 $\pm$ 0.30 & \textbf{12.19 $\pm$ 0.17}& 13.02 $\pm$ 0.26 & \textbf{11.48 $\pm$ 0.18}& 16.14 $\pm$ 0.31 & \textbf{14.40 $\pm$ 0.20}& 0.086 $\pm$ 0.035 & \textbf{0.271 $\pm$ 0.020}\\
			GEM & 12.86 $\pm$ 0.29 & \textbf{10.86 $\pm$  0.26}& 12.96 $\pm$ 0.15 & \textbf{11.06 $\pm$  0.27}& 13.05 $\pm$ 0.11 & 11.42 $\pm$  0.27& 13.25 $\pm$ 0.10 & 11.84 $\pm$  0.34& 13.53 $\pm$ 0.09 & 12.38 $\pm$  0.38& 13.13 $\pm$ 0.12 & 11.52 $\pm$  0.30& 16.26 $\pm$ 0.09 & 14.45 $\pm$ 0.37& 0.073 $\pm$ 0.011 & 0.266 $\pm$ 0.038\\
			\bottomrule
		\end{tabular}%
	}
\end{table*}

\subsection{Supplementary AUPRC results}

\textbf{CODE-II sex classification.} Among models pretrained on PTB-XL paper ECGs, RobECG-CL achieves the highest average AUPRC under both training fractions, as shown in Table~\ref{tab:codeii-sex-auprc}. Although MoCo performs better on clean paper ECGs, its performance decreases under progressive degradation. From clean paper ECGs to Severity~4, the AUPRC of RobECG-CL decreases by only 0.024 and 0.034 under the 1\% and 10\% settings; by contrast, it is 0.128 and 0.118 for MoCo. Additionally, RobECG-CL also outperforms PULSE, GEM, and the waveform foundation models in average AUPRC under the 1\% setting.

\textbf{EchoNext SHD classification.} RobECG-CL also achieves the highest average AUPRC among the PTB-XL-pretrained methods on EchoNext (Table~\ref{tab:echonext-auprc}). From clean paper ECGs to Severity~4, RobECG-CL shows AUPRC decreases of 0.023 and 0.029, whereas the corresponding decreases for MoCo are 0.135 and 0.143. PULSE and GEM exhibit higher average AUPRC but larger degradation-related decreases. These results suggest that RobECG-CL preserves strong robustness under progressive paper ECG degradation.

\begin{table*}[h]
	\centering
	\scriptsize
	\setlength{\tabcolsep}{2.4pt}
	\caption{AUPRC across degradation severity levels on CODE-II sex classification. Higher severity levels denote progressively stronger degradation. Values are mean $\pm$ standard deviation over five runs.}
	\label{tab:codeii-sex-auprc}
	\resizebox{\textwidth}{!}{%
		\begin{tabular}{lcccccccccccc}
			\toprule
			\multirow{2}{*}{Method} & \multicolumn{2}{c}{Clean Paper ECG} & \multicolumn{2}{c}{Severity 1} & \multicolumn{2}{c}{Severity 2} & \multicolumn{2}{c}{Severity 3} & \multicolumn{2}{c}{Severity 4} & \multicolumn{2}{c}{Avg.} \\
			\cmidrule(lr){2-3}\cmidrule(lr){4-5}\cmidrule(lr){6-7}\cmidrule(lr){8-9}\cmidrule(lr){10-11}\cmidrule(lr){12-13}
			& 1\% & 10\% & 1\% & 10\% & 1\% & 10\% & 1\% & 10\% & 1\% & 10\% & 1\% & 10\% \\
			\midrule
			\multicolumn{13}{l}{\emph{PTB-XL $\rightarrow$ downstream}} \\
			RobECG-CL & 0.695 $\pm$ 0.006 & 0.739 $\pm$ 0.007 & \textbf{0.695 $\pm$ 0.005} & 0.736 $\pm$ 0.006 & \textbf{0.692 $\pm$ 0.006} & \textbf{0.731 $\pm$ 0.006} & \textbf{0.688 $\pm$ 0.006} & \textbf{0.725 $\pm$ 0.005} & \textbf{0.671 $\pm$ 0.005} & \textbf{0.705 $\pm$ 0.005} & \textbf{0.688 $\pm$ 0.005} & \textbf{0.727 $\pm$ 0.006} \\
			MoCo & \textbf{0.696 $\pm$ 0.014} & \textbf{0.755 $\pm$ 0.005} & 0.678 $\pm$ 0.012 & \textbf{0.744 $\pm$ 0.005} & 0.650 $\pm$ 0.012 & 0.717 $\pm$ 0.006 & 0.611 $\pm$ 0.014 & 0.680 $\pm$ 0.006 & 0.568 $\pm$ 0.015 & 0.637 $\pm$ 0.005 & 0.645 $\pm$ 0.011 & 0.710 $\pm$ 0.004 \\
			DINO & 0.648 $\pm$ 0.014 & 0.722 $\pm$ 0.007 & 0.647 $\pm$ 0.013 & 0.720 $\pm$ 0.005 & 0.622 $\pm$ 0.016 & 0.707 $\pm$ 0.003 & 0.583 $\pm$ 0.018 & 0.680 $\pm$ 0.003 & 0.541 $\pm$ 0.016 & 0.640 $\pm$ 0.003 & 0.612 $\pm$ 0.013 & 0.696 $\pm$ 0.004 \\
			BYOL & 0.665 $\pm$ 0.008 & 0.745 $\pm$ 0.004 & 0.650 $\pm$ 0.006 & 0.736 $\pm$ 0.004 & 0.624 $\pm$ 0.006 & 0.713 $\pm$ 0.004 & 0.590 $\pm$ 0.009 & 0.680 $\pm$ 0.003 & 0.549 $\pm$ 0.011 & 0.640 $\pm$ 0.003 & 0.619 $\pm$ 0.005 & 0.706 $\pm$ 0.003 \\
			SimCLR & 0.652 $\pm$ 0.009 & 0.695 $\pm$ 0.003 & 0.644 $\pm$ 0.008 & 0.691 $\pm$ 0.004 & 0.623 $\pm$ 0.008 & 0.680 $\pm$ 0.005 & 0.592 $\pm$ 0.008 & 0.659 $\pm$ 0.005 & 0.554 $\pm$ 0.009 & 0.623 $\pm$ 0.005 & 0.615 $\pm$ 0.007 & 0.671 $\pm$ 0.003 \\
			\multicolumn{13}{l}{\emph{Pretrained (image-based) $\rightarrow$ downstream}} \\
			PULSE & \textbf{0.709 $\pm$ 0.023} & 0.774 $\pm$  0.005& \textbf{0.705 $\pm$ 0.021} & 0.766 $\pm$  0.008& \textbf{0.692 $\pm$ 0.022} & 0.749 $\pm$  0.011& \textbf{0.666 $\pm$ 0.021} & 0.725 $\pm$  0.010& 0.627 $\pm$ 0.021 & 0.692 $\pm$  0.010& \textbf{0.681 $\pm$ 0.019} & 0.742 $\pm$  0.009\\
			GEM & 0.708 $\pm$ 0.021 & \textbf{0.776 $\pm$  0.010}& 0.700 $\pm$ 0.024 & \textbf{0.770 $\pm$  0.009}& 0.686 $\pm$ 0.023 & \textbf{0.756 $\pm$  0.008}& 0.663 $\pm$ 0.023 & \textbf{0.734 $\pm$  0.007}& \textbf{0.630 $\pm$ 0.021} & \textbf{0.701 $\pm$  0.008}& 0.678 $\pm$ 0.021 & \textbf{0.748 $\pm$  0.008}\\
			\multicolumn{13}{l}{\emph{Pretrained (waveform-based) $\rightarrow$ downstream}} \\
			CSFM & -- & -- & -- & -- & -- & -- & -- & -- & -- & -- & \textbf{0.676 $\pm$ 0.044} & \textbf{0.827 $\pm$ 0.004} \\
			ECGFounder & -- & -- & -- & -- & -- & -- & -- & -- & -- & -- & 0.543 $\pm$ 0.036 & 0.813 $\pm$ 0.009 \\
			ECG\_FM & -- & -- & -- & -- & -- & -- & -- & -- & -- & -- & 0.602 $\pm$ 0.045 & 0.820 $\pm$ 0.006 \\
			\bottomrule
		\end{tabular}%
	}
\end{table*}

\begin{table*}[h]
	\centering
	\scriptsize
	\setlength{\tabcolsep}{2.4pt}
	\caption{AUPRC across degradation severity levels on EchoNext structural heart disease classification. Higher severity levels denote progressively stronger degradation. Values are mean $\pm$ standard deviation over five runs. For baseline methods, severity-specific AUPRC is not available in the source table and is marked as missing.}
	\label{tab:echonext-auprc}
	\resizebox{\textwidth}{!}{%
		\begin{tabular}{lcccccccccccc}
			\toprule
			\multirow{2}{*}{Method} & \multicolumn{2}{c}{Clean Paper ECG} & \multicolumn{2}{c}{Severity 1} & \multicolumn{2}{c}{Severity 2} & \multicolumn{2}{c}{Severity 3} & \multicolumn{2}{c}{Severity 4} & \multicolumn{2}{c}{Avg.} \\
			\cmidrule(lr){2-3}\cmidrule(lr){4-5}\cmidrule(lr){6-7}\cmidrule(lr){8-9}\cmidrule(lr){10-11}\cmidrule(lr){12-13}
			& 1\% & 10\% & 1\% & 10\% & 1\% & 10\% & 1\% & 10\% & 1\% & 10\% & 1\% & 10\% \\
			\midrule
			\multicolumn{13}{l}{\emph{PTB-XL $\rightarrow$ downstream}} \\
			RobECG-CL & 0.624 $\pm$ 0.008& 0.683 $\pm$ 0.003& 0.622 $\pm$ 0.009& 0.679 $\pm$ 0.003& \textbf{0.621 $\pm$ 0.009}& \textbf{0.679 $\pm$ 0.004} & \textbf{0.617 $\pm$ 0.010} & \textbf{0.673 $\pm$ 0.003} & \textbf{0.601 $\pm$ 0.009} & \textbf{0.654 $\pm$ 0.004} & \textbf{0.617 $\pm$ 0.009} & \textbf{0.674 $\pm$ 0.003} \\
			MoCo & \textbf{0.663 $\pm$  0.004}& \textbf{0.715 $\pm$  0.005}& \textbf{0.644 $\pm$  0.005}& 0.698 $\pm$  0.005& 0.611 $\pm$  0.007& 0.665 $\pm$  0.007& 0.571 $\pm$  0.014& 0.626 $\pm$  0.006& 0.528 $\pm$  0.011& 0.572 $\pm$  0.006& 0.609 $\pm$ 0.005 & 0.661 $\pm$ 0.005 \\
			DINO & 0.650 $\pm$  0.006& 0.712 $\pm$  0.003& 0.628 $\pm$  0.007& \textbf{0.701 $\pm$  0.003}& 0.600 $\pm$  0.013& 0.679 $\pm$  0.003& 0.555 $\pm$  0.017& 0.636 $\pm$  0.005& 0.520 $\pm$  0.016& 0.594 $\pm$  0.005& 0.595 $\pm$ 0.010 & 0.669 $\pm$ 0.003 \\
			BYOL & 0.654 $\pm$  0.005& 0.709 $\pm$  0.003& 0.624 $\pm$  0.008& 0.690 $\pm$  0.003& 0.597 $\pm$  0.009& 0.665 $\pm$  0.004& 0.558 $\pm$  0.014& 0.631 $\pm$  0.005& 0.516 $\pm$  0.013& 0.588 $\pm$  0.005& 0.593 $\pm$ 0.007 & 0.660 $\pm$ 0.003 \\
			SimCLR & 0.577 $\pm$  0.015& 0.613 $\pm$  0.004& 0.568 $\pm$  0.013& 0.604 $\pm$  0.005& 0.553 $\pm$  0.014& 0.590 $\pm$  0.004& 0.533 $\pm$  0.012& 0.564 $\pm$  0.006& 0.498 $\pm$  0.005& 0.528 $\pm$  0.005& 0.546 $\pm$ 0.013 & 0.581 $\pm$ 0.004 \\
			\multicolumn{13}{l}{\emph{Pretrained (image-based) $\rightarrow$ downstream}} \\
			PULSE & \textbf{0.708 $\pm$  0.011}& 0.747 $\pm$  0.008& \textbf{0.689 $\pm$  0.017}& 0.740 $\pm$  0.007& \textbf{0.668 $\pm$  0.016}& 0.726 $\pm$  0.005& \textbf{0.642 $\pm$  0.021}& 0.705 $\pm$  0.005& \textbf{0.607 $\pm$  0.019}& \textbf{0.668 $\pm$  0.010}& \textbf{0.664 $\pm$ 0.016} & 0.717 $\pm$  0.005\\
			GEM & 0.692 $\pm$  0.019& \textbf{0.748 $\pm$  0.010}& 0.673 $\pm$  0.021& \textbf{0.741 $\pm$  0.010}& 0.656 $\pm$  0.017& \textbf{0.727 $\pm$  0.013}& 0.635 $\pm$  0.017& \textbf{0.705 $\pm$  0.015}& 0.594 $\pm$  0.024& 0.663 $\pm$  0.018& 0.651 $\pm$ 0.018 & \textbf{0.717 $\pm$  0.014}\\
			\multicolumn{13}{l}{\emph{Pretrained (waveform-based) $\rightarrow$ downstream}} \\
			CSFM & -- & -- & -- & -- & -- & -- & -- & -- & -- & -- & 0.150 $\pm$ 0.037 & 0.238 $\pm$ 0.003 \\
			ECGFounder & -- & -- & -- & -- & -- & -- & -- & -- & -- & -- & 0.125 $\pm$ 0.018 & 0.215 $\pm$ 0.007 \\
			ECG\_FM & -- & -- & -- & -- & -- & -- & -- & -- & -- & -- & \textbf{0.173 $\pm$ 0.052} & \textbf{0.272 $\pm$ 0.007} \\
			\bottomrule
		\end{tabular}%
	}
\end{table*}

\subsection{Supplementary real-world paper ECG results}
We further evaluate all methods on 312 real-world paper ECGs including 37 diagnostic labels. All results are pooled out-of-fold predictions obtained using five-fold evaluation.

\textbf{Performance across all diagnostic labels.} RobECG-CL achieves the best average macro AUROC across all 37 diagnostic labels. RobECG-CL also remains comparable to the strongest baselines in macro AUPRC and F1, compared with MoCo and GEM, and shows strong performance for several individual labels (Table~\ref{tab:real-world-hospital-all37}). Specifically, RobECG-CL performs an AUROC/AUPRC/F1 of 0.928/0.724/0.691 for complete right bundle branch block (CRBBB), 0.912/0.552/0.635 for sinus bradycardia (SB), and 0.841/0.404/0.299 for normal ECGs (NORM). Although the best-performing model varies across different labels, RobECG-CL demonstrates the most consistent overall discrimination across the entire diagnostic label space.

\begin{table*}[h]
	\centering
	\scriptsize
	\setlength{\tabcolsep}{1.6pt}
	\caption{Per-label AUROC, AUPRC, and F1 on all 37 diagnostic labels in the real-world paper ECG set collected from The First Affiliated Hospital with Nanjing Medical University. Results are pooled out-of-fold performance under strict patient-aware 5-fold evaluation.}
	\label{tab:real-world-hospital-all37}
	\resizebox{\textwidth}{!}{%
		\begin{tabular}{lccccccccccccccccccccc}
			\toprule
			\multirow{2}{*}{Label} & \multicolumn{3}{c}{RobECG-CL} & \multicolumn{3}{c}{MoCo} & \multicolumn{3}{c}{DINO} & \multicolumn{3}{c}{BYOL} & \multicolumn{3}{c}{SimCLR} & \multicolumn{3}{c}{PULSE} & \multicolumn{3}{c}{GEM} \\
			\cmidrule(lr){2-4}\cmidrule(lr){5-7}\cmidrule(lr){8-10}\cmidrule(lr){11-13}\cmidrule(lr){14-16}\cmidrule(lr){17-19}\cmidrule(lr){20-22}
			& AUROC & AUPRC & F1 & AUROC & AUPRC & F1 & AUROC & AUPRC & F1 & AUROC & AUPRC & F1 & AUROC & AUPRC & F1 & AUROC & AUPRC & F1 & AUROC & AUPRC & F1 \\
			\midrule
			1AVB (36) & \textbf{0.581} & 0.149 & \textbf{0.212} & \underline{0.541} & 0.158 & \underline{0.171} & 0.499 & 0.124 & 0.171 & 0.500 & \underline{0.167} & 0.134 & 0.510 & 0.129 & 0.148 & 0.531 & 0.136 & 0.131 & 0.532 & \textbf{0.176} & 0.140 \\
			AF (28) & 0.753 & 0.278 & 0.278 & 0.752 & \underline{0.325} & \textbf{0.393} & \underline{0.778} & 0.309 & \underline{0.295} & 0.617 & 0.205 & 0.273 & 0.618 & 0.220 & 0.256 & 0.750 & 0.264 & 0.275 & \textbf{0.818} & \textbf{0.385} & 0.244 \\
			CRBBB (36) & \textbf{0.928} & \textbf{0.724} & \textbf{0.691} & 0.877 & 0.656 & 0.597 & 0.880 & 0.581 & 0.571 & 0.848 & 0.563 & 0.522 & 0.722 & 0.248 & 0.278 & 0.895 & 0.676 & 0.597 & \underline{0.908} & \underline{0.693} & \underline{0.677} \\
			IRBBB (24) & \underline{0.626} & 0.157 & 0.140 & \textbf{0.645} & \textbf{0.259} & 0.192 & 0.604 & \underline{0.209} & \textbf{0.213} & 0.574 & 0.139 & 0.139 & 0.540 & 0.128 & 0.136 & 0.569 & 0.159 & 0.162 & 0.607 & 0.198 & \underline{0.194} \\
			ISC (41) & 0.653 & 0.240 & 0.271 & 0.736 & 0.249 & \underline{0.346} & \underline{0.750} & \underline{0.317} & 0.293 & 0.657 & 0.211 & 0.259 & \textbf{0.788} & \textbf{0.428} & \textbf{0.351} & 0.688 & 0.246 & 0.205 & 0.726 & 0.264 & 0.270 \\
			LAE (20) & \textbf{0.830} & 0.246 & \textbf{0.429} & 0.777 & 0.259 & 0.226 & 0.702 & 0.146 & 0.222 & 0.745 & \underline{0.268} & 0.203 & \underline{0.790} & 0.200 & 0.189 & 0.737 & 0.268 & 0.138 & 0.721 & \textbf{0.309} & \underline{0.231} \\
			LVH (23) & \underline{0.952} & \textbf{0.699} & \underline{0.615} & 0.944 & 0.572 & 0.516 & 0.877 & 0.447 & 0.324 & \textbf{0.960} & \underline{0.655} & \textbf{0.636} & 0.928 & 0.604 & 0.415 & 0.859 & 0.556 & 0.579 & 0.816 & 0.512 & 0.474 \\
			NORM (23) & \textbf{0.841} & \textbf{0.404} & \textbf{0.299} & 0.755 & \underline{0.322} & 0.246 & \underline{0.839} & 0.311 & \underline{0.293} & 0.792 & 0.258 & 0.250 & 0.778 & 0.224 & 0.233 & 0.774 & 0.230 & 0.222 & 0.791 & 0.235 & 0.118 \\
			OMI (26) & 0.557 & 0.104 & 0.177 & \underline{0.628} & \underline{0.189} & \underline{0.186} & \textbf{0.628} & \textbf{0.225} & \textbf{0.212} & 0.575 & 0.112 & 0.115 & 0.598 & 0.161 & 0.182 & 0.493 & 0.127 & 0.057 & 0.514 & 0.165 & 0.179 \\
			PAC (20) & 0.699 & 0.185 & \textbf{0.219} & \underline{0.764} & 0.160 & 0.156 & \textbf{0.799} & \underline{0.194} & 0.203 & 0.729 & 0.144 & 0.216 & 0.691 & \textbf{0.241} & \underline{0.216} & 0.635 & 0.102 & 0.000 & 0.635 & 0.115 & 0.138 \\
			PVC (41) & 0.842 & 0.671 & 0.590 & 0.899 & 0.661 & 0.548 & 0.903 & 0.621 & 0.635 & 0.884 & 0.612 & 0.577 & 0.813 & 0.436 & 0.463 & \textbf{0.915} & \textbf{0.754} & \textbf{0.685} & \underline{0.908} & \underline{0.727} & \underline{0.683} \\
			SB (42) & \textbf{0.912} & \underline{0.552} & \textbf{0.635} & \underline{0.892} & 0.545 & \underline{0.558} & 0.848 & 0.423 & 0.467 & 0.889 & 0.467 & 0.454 & 0.840 & \textbf{0.553} & 0.500 & 0.816 & 0.392 & 0.314 & 0.811 & 0.414 & 0.406 \\
			2AVB (1) & 0.707 & 0.011 & 0.000 & \underline{0.714} & \underline{0.011} & 0.000 & \textbf{0.891} & \textbf{0.029} & 0.000 & 0.637 & 0.009 & 0.000 & 0.492 & 0.006 & 0.000 & 0.248 & 0.004 & 0.000 & 0.595 & 0.008 & 0.000 \\
			3AVB (3) & \underline{0.963} & \underline{0.177} & 0.000 & 0.810 & 0.035 & 0.000 & 0.926 & 0.078 & 0.000 & \textbf{0.965} & \textbf{0.278} & \textbf{0.444} & 0.538 & 0.022 & 0.000 & 0.817 & 0.040 & 0.000 & 0.946 & 0.131 & 0.000 \\
			AFL (6) & 0.551 & 0.045 & 0.065 & 0.760 & \underline{0.225} & 0.085 & \underline{0.836} & 0.116 & \textbf{0.105} & \textbf{0.849} & \textbf{0.265} & \underline{0.088} & 0.789 & 0.068 & 0.041 & 0.813 & 0.147 & 0.000 & 0.817 & 0.177 & 0.000 \\
			AMI (8) & \underline{0.781} & 0.206 & 0.150 & 0.726 & 0.206 & 0.113 & 0.710 & 0.161 & 0.109 & 0.690 & 0.125 & 0.092 & 0.640 & 0.049 & 0.066 & 0.771 & \underline{0.231} & \underline{0.182} & \textbf{0.785} & \textbf{0.393} & \textbf{0.400} \\
			ASMI (10) & 0.724 & 0.189 & 0.070 & \underline{0.846} & \textbf{0.394} & \underline{0.244} & \textbf{0.884} & 0.330 & \textbf{0.320} & 0.840 & \underline{0.348} & 0.207 & 0.710 & 0.219 & 0.126 & 0.820 & 0.324 & 0.167 & 0.800 & 0.283 & 0.182 \\
			AT (2) & \textbf{0.731} & \underline{0.029} & 0.000 & 0.656 & 0.013 & 0.000 & 0.255 & 0.006 & 0.000 & 0.592 & 0.011 & 0.000 & 0.642 & 0.013 & 0.000 & \underline{0.703} & 0.017 & 0.000 & 0.581 & \textbf{0.059} & 0.000 \\
			CLBBB (5) & \textbf{0.857} & \underline{0.148} & \textbf{0.081} & \underline{0.851} & 0.086 & 0.064 & 0.776 & 0.043 & 0.042 & 0.850 & \textbf{0.171} & \underline{0.073} & 0.484 & 0.019 & 0.038 & 0.711 & 0.058 & 0.000 & 0.771 & 0.081 & 0.000 \\
			EAMI (10) & 0.803 & 0.219 & 0.167 & 0.845 & 0.367 & \underline{0.316} & 0.708 & 0.213 & 0.207 & 0.675 & 0.312 & 0.081 & 0.655 & 0.171 & 0.091 & \textbf{0.858} & \underline{0.429} & \textbf{0.462} & \underline{0.856} & \textbf{0.477} & \textbf{0.462} \\
			ILBBB (1) & 0.563 & 0.007 & 0.000 & \underline{0.685} & \underline{0.010} & 0.000 & 0.064 & 0.003 & 0.000 & 0.289 & 0.005 & 0.000 & 0.148 & 0.004 & 0.000 & \textbf{0.846} & \textbf{0.020} & 0.000 & 0.129 & 0.004 & 0.000 \\
			IMI (5) & 0.560 & 0.034 & 0.031 & \textbf{0.791} & 0.058 & \underline{0.054} & \underline{0.745} & \textbf{0.074} & \textbf{0.066} & 0.643 & 0.028 & 0.038 & 0.582 & 0.035 & 0.019 & 0.619 & \underline{0.069} & 0.000 & 0.615 & 0.027 & 0.000 \\
			IVCD (13) & 0.522 & 0.057 & 0.078 & 0.556 & 0.051 & \underline{0.079} & 0.552 & 0.047 & \textbf{0.083} & 0.347 & 0.038 & 0.054 & 0.426 & 0.036 & 0.063 & \textbf{0.631} & \textbf{0.106} & 0.000 & \underline{0.606} & \underline{0.095} & 0.000 \\
			JE (2) & 0.981 & 0.196 & 0.000 & \textbf{0.992} & \underline{0.417} & \textbf{0.286} & 0.898 & 0.044 & 0.000 & \underline{0.990} & \textbf{0.625} & \underline{0.222} & 0.903 & 0.074 & 0.000 & 0.947 & 0.079 & 0.000 & 0.897 & 0.068 & 0.000 \\
			LAFB (13) & \textbf{0.932} & \underline{0.433} & 0.364 & 0.799 & 0.241 & 0.222 & 0.771 & 0.194 & 0.167 & 0.759 & 0.159 & 0.270 & 0.849 & 0.255 & 0.125 & \underline{0.878} & 0.405 & \underline{0.381} & 0.865 & \textbf{0.470} & \textbf{0.455} \\
			LNGQT (6) & \textbf{0.665} & 0.041 & \textbf{0.039} & 0.499 & 0.021 & 0.022 & 0.447 & 0.022 & \underline{0.029} & 0.533 & 0.023 & 0.025 & 0.572 & 0.041 & 0.015 & \underline{0.629} & \underline{0.048} & 0.000 & 0.570 & \textbf{0.070} & 0.000 \\
			PACE-A (1) & \underline{0.370} & \underline{0.005} & 0.000 & 0.113 & 0.004 & 0.000 & \textbf{0.826} & \textbf{0.018} & 0.000 & 0.318 & 0.005 & 0.000 & 0.158 & 0.004 & 0.000 & 0.077 & 0.003 & 0.000 & 0.087 & 0.004 & 0.000 \\
			PACE-D (5) & \textbf{0.747} & \textbf{0.069} & \underline{0.050} & 0.639 & 0.028 & 0.023 & 0.352 & 0.017 & 0.015 & 0.494 & 0.020 & 0.038 & 0.609 & \underline{0.049} & \textbf{0.087} & \underline{0.730} & 0.041 & 0.000 & 0.694 & 0.033 & 0.000 \\
			PACE-V (12) & 0.877 & 0.342 & 0.296 & \textbf{0.923} & \textbf{0.444} & \textbf{0.429} & 0.747 & 0.163 & 0.093 & \underline{0.897} & \underline{0.364} & 0.324 & 0.878 & 0.319 & 0.316 & 0.778 & 0.285 & \underline{0.381} & 0.779 & 0.258 & 0.267 \\
			RVH (4) & 0.732 & 0.078 & 0.031 & \underline{0.839} & 0.061 & \textbf{0.070} & 0.635 & 0.037 & 0.028 & 0.592 & 0.039 & \underline{0.042} & 0.209 & 0.010 & 0.016 & 0.727 & \textbf{0.105} & 0.000 & \textbf{0.867} & \underline{0.080} & 0.000 \\
			SA (1) & 0.354 & 0.005 & 0.000 & 0.486 & 0.006 & 0.000 & 0.367 & 0.005 & 0.000 & \textbf{0.958} & \textbf{0.071} & 0.000 & \underline{0.704} & \underline{0.011} & 0.000 & 0.391 & 0.005 & 0.000 & 0.272 & 0.004 & 0.000 \\
			SArrh (4) & 0.304 & 0.011 & 0.013 & 0.211 & 0.010 & 0.011 & 0.250 & 0.010 & \textbf{0.019} & 0.189 & 0.009 & 0.000 & \textbf{0.506} & \textbf{0.016} & \underline{0.016} & \underline{0.421} & \underline{0.014} & 0.000 & 0.312 & 0.011 & 0.000 \\
			STach (10) & 0.919 & 0.369 & 0.364 & 0.945 & \underline{0.411} & 0.357 & \textbf{0.976} & 0.408 & \textbf{0.600} & 0.949 & \textbf{0.458} & 0.429 & 0.919 & 0.342 & 0.421 & \underline{0.959} & 0.370 & \underline{0.476} & 0.951 & 0.363 & 0.273 \\
			STTC (13) & 0.584 & 0.061 & 0.089 & 0.645 & \underline{0.082} & \underline{0.095} & \underline{0.698} & 0.077 & 0.094 & 0.545 & 0.052 & 0.066 & \textbf{0.757} & \textbf{0.114} & \textbf{0.128} & 0.629 & 0.059 & 0.000 & 0.587 & 0.062 & 0.000 \\
			VE (1) & \textbf{0.550} & \textbf{0.007} & 0.000 & 0.222 & 0.004 & 0.000 & \underline{0.424} & \underline{0.006} & 0.000 & 0.151 & 0.004 & 0.000 & 0.013 & 0.003 & 0.000 & 0.106 & 0.004 & 0.000 & 0.167 & 0.004 & 0.000 \\
			VT (3) & \textbf{0.865} & 0.359 & 0.286 & 0.725 & 0.347 & 0.222 & \underline{0.824} & 0.138 & 0.000 & 0.783 & \underline{0.363} & \underline{0.400} & 0.574 & 0.342 & 0.222 & 0.775 & 0.354 & 0.000 & 0.761 & \textbf{0.389} & \textbf{0.500} \\
			WPW (5) & \underline{0.762} & 0.108 & 0.154 & \textbf{0.767} & 0.146 & 0.087 & 0.759 & 0.079 & \underline{0.211} & 0.622 & \underline{0.146} & 0.025 & 0.648 & 0.095 & 0.121 & 0.696 & \textbf{0.251} & \textbf{0.333} & 0.731 & 0.086 & 0.000 \\
			\midrule
			Avg. & \textbf{0.718} & 0.206 & \underline{0.186} & \underline{0.710} & \textbf{0.217} & \textbf{0.187} & 0.687 & 0.168 & 0.165 & 0.674 & 0.209 & 0.181 & 0.622 & 0.159 & 0.143 & 0.682 & 0.199 & 0.155 & 0.671 & \underline{0.212} & 0.170 \\
			\bottomrule
		\end{tabular}%
	}
\end{table*}

\begin{figure*}[t!]
    \centering
    \includegraphics[width=\textwidth]
    {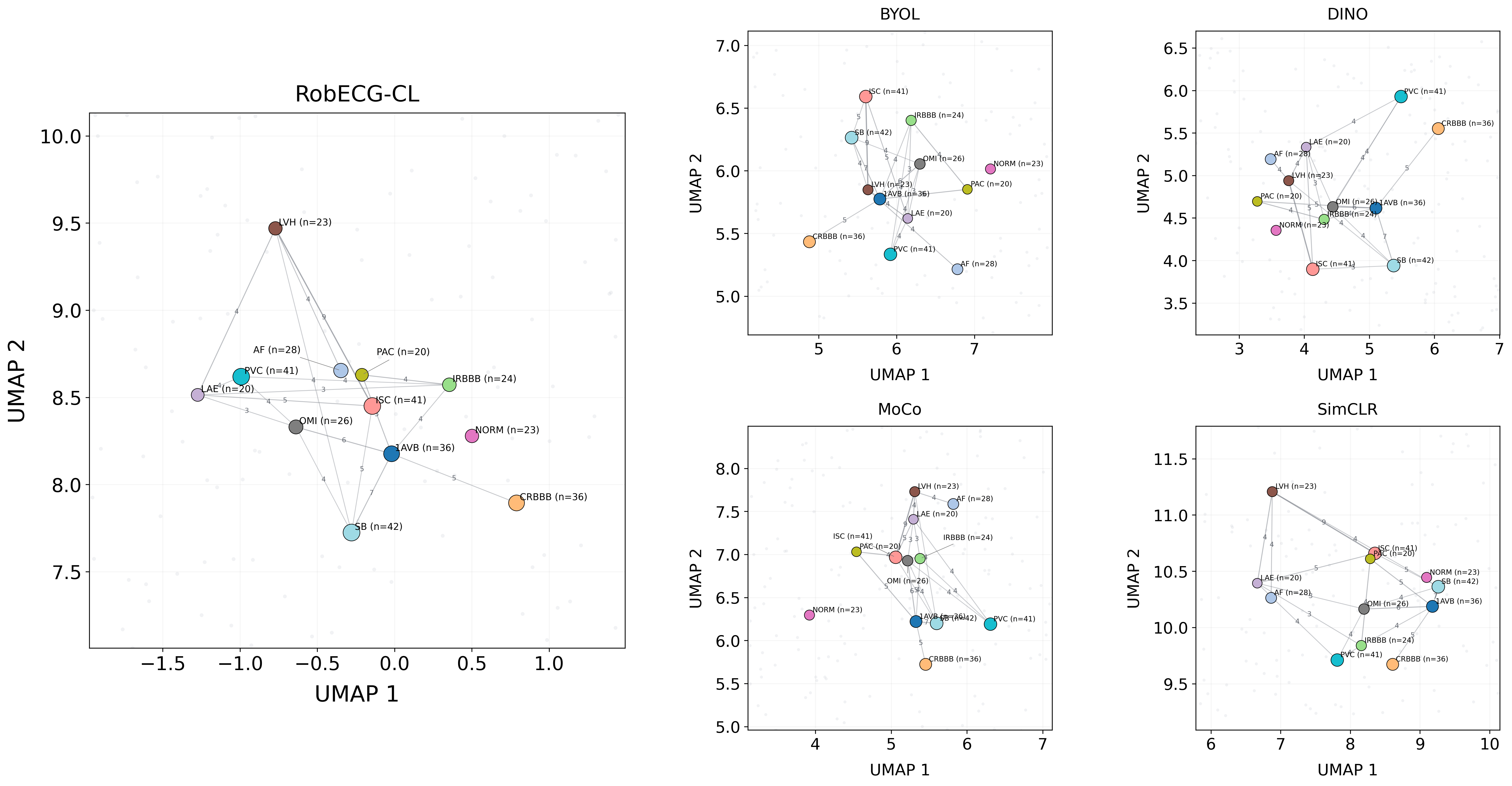}
    \caption{Diagnostic-label centroids in the semantic UMAP spaces of RobECG-CL, BYOL, DINO, MoCo, and SimCLR. Here, \(n\) denotes the number of ECGs positive for each label. For each method, the standardized per-label logits are projected into a two-dimensional UMAP space. Each colored point is then calculated as the arithmetic mean of the UMAP coordinates of all ECGs positive for the corresponding label. Edges connect label pairs with the strongest true-label co-occurrence, while the number on each edge indicates the number of ECGs positive for both labels.}
    \label{fig:real-world-centroids}
\end{figure*}

\textbf{The real-world semantic prediction space.} Figure~\ref{fig:real-world-centroids} visualizes the diagnostic-label centroids calculated from the pooled out-of-fold predictions. For RobECG-CL, CRBBB, SB, and LVH are clearly separable, resulting in strong classification performance.
The left ventricular hypertrophy (LVH) and sinus bradycardia (SB) separated from other diagnostic labels. Atrial fibrillation (AF) and premature atrial contractions (PAC) located close to each other. From a clinical perspective, atrial arrhythmias occur a continuum from premature atrial contractions to flutter and fibrillation, and these rhythm abnormalities often coexist. 
The centroid distance of PVC and left atrial enlargement (LAE) were close either. Clinically, patients with left ventricular dysfunction may present with higher PVC burden and left atrial enlargement. Such coexistence of multiple cardiac abnormalities may reduce the model's ability to distinguish ECG categories as clearly as in cases with a single diagnostic label.

\textbf{Impact of label imbalance.}
The complete 37-label evaluation includes several extremely rare samples. In particular, 2AVB, ILBBB, PACE-A, SA, and VE contain only one positive sample, while 3AVB, AT, JE, CLBBB, IMI, PACE-D, RVH, SArrh, VT, and WPW contain only two to five positive samples. Therefore, results for these labels are sensitive to indivisual samples,  and high AUROC values may co-exist with low AUPRC or zero F1.

\begin{figure*}[h]
    \centering
    \includegraphics[width=\textwidth]
    {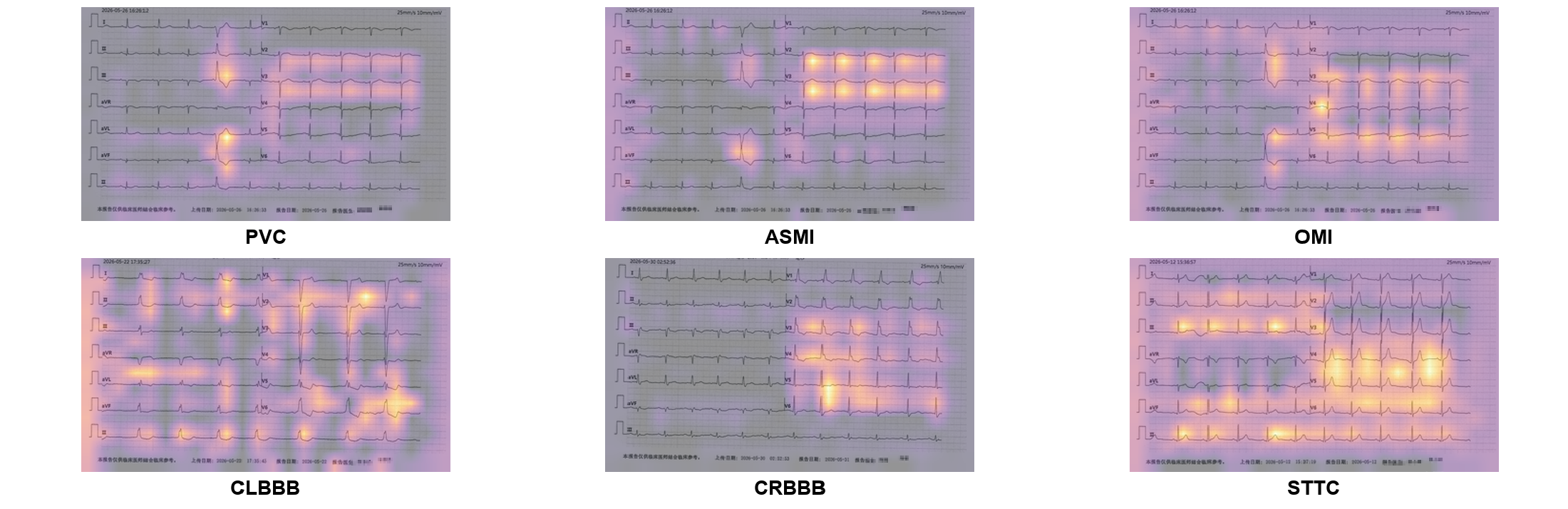}
    \caption{Grad-CAM heatmap visualizations on real-world paper ECGs. Each panel shows the attribution map for one diagnostic label from the external hospital dataset. RobECG-CL highlights localized ECG regions across diverse diagnoses, suggesting that the learned representation remains interpretable under real-world paper ECG degradation.}
    \label{fig:realworld-heatmap}
\end{figure*}

\textbf{Grad-CAM.}
Figure~\ref{fig:realworld-heatmap} shows representative Grad-CAM visualizations on real-world paper ECGs. Across different diagnostic labels, RobECG-CL attends to localized ECG regions rather than background artifacts.
The highest interpretability was observed for paced rhythm, premature ventricular contraction (PVC), and left ventricular hypertrophy (LVH). The first row of Figure 4 derived from a single patient with multiple diagnostic labels. Although PVC, ASMI, and OMI were all identified accurately, the presence of multiple labels resulted in interactions among different diagnostic features, leading to overlapping and mixed highlighted regions. 
The second row of Figure 4 was derived from three other patients. The visualization for CLBBB was clearly more accurate than that for CRBBB. For CRBBB, the V1 and V2 leads were not highlighted, although it's relatively easy to recognize in routine ECG interpretation. 
The highlighted regions for STTC showed no clear specificity. Overall, the Grad-CAM maps indicate that the model mainly focused on the precordial leads rather than the limb leads, with the highlighted regions mainly focused on QRS complexes.

\section{Synthetic dataset curation}
Our synthetic ECG dataset comprises four progressive degradation stages. The dataset maintains a balanced distribution across both lead layouts (50\% $2 \times 6$ with red grids, 50\% $4 \times 3$ with black-and-white grids) and segmentation strategies (50\% synchronous, 50\% asynchronous).

Clean paper ECGs (Severity 0) consist of $224 \times 224$ images rendered directly from raw WFDB signals. For Severities 1 to 4, we employ a cumulative degradation strategy that progressively injects realistic artifacts encompassing rotation, color temperature shifts, cropping, noise, Gaussian blur, and wrinkling. 
The degradation severity monotonically increases at each stage. Specifically, the continuous parameters inherited from the previous stage are incremented by sampling from Gaussian distributions.  
A comprehensive summary of the parameter configurations, their physical interpretations, and corresponding implementations is detailed in Table~\ref{tab:degradation}. 

\begin{table*}[h]
\centering
\caption{Summary of progressive degradation methods, parameter step distributions, physical interpretations, and corresponding implementations.}
\label{tab:degradation}
\resizebox{\textwidth}{!}{%
\begin{tabular}{llll}
\toprule
\textbf{Method} & \textbf{Step Distribution ($\epsilon$)} & \textbf{Physical Interpretation} & \textbf{Implementation} \\
\midrule
Rotation & $\mathcal{N}(0, 5)$ & Paper misalignment during scanning/photography. & \texttt{iaa.Affine} \\
Color Temperature & $\mathcal{N}(0, 300)$ & Variations in white-balance and ambient illumination. & \texttt{iaa.ChangeColorTemperature} \\
Cropping & $\mathcal{N}(0.002, 0.01)$ & Partial border loss or automatic edge trimming. & \texttt{iaa.Crop} \\
Noise & $\mathcal{N}(2, 6)$ & Pixel-level noise introduced by camera/scanner sensors. & \texttt{iaa.AdditiveGaussianNoise} \\
Gaussian Blur & $\mathcal{N}(1, 3)$ & Defocus, motion blur, and image-softening effects. & \texttt{cv2.GaussianBlur} \\
\bottomrule
\end{tabular}%
}
\end{table*}

\section{Real-world dataset curation}
\label{sec:appendix_real_world_data}
\begin{figure*}[t]
    \centering
    \includegraphics[width=0.8\textwidth]{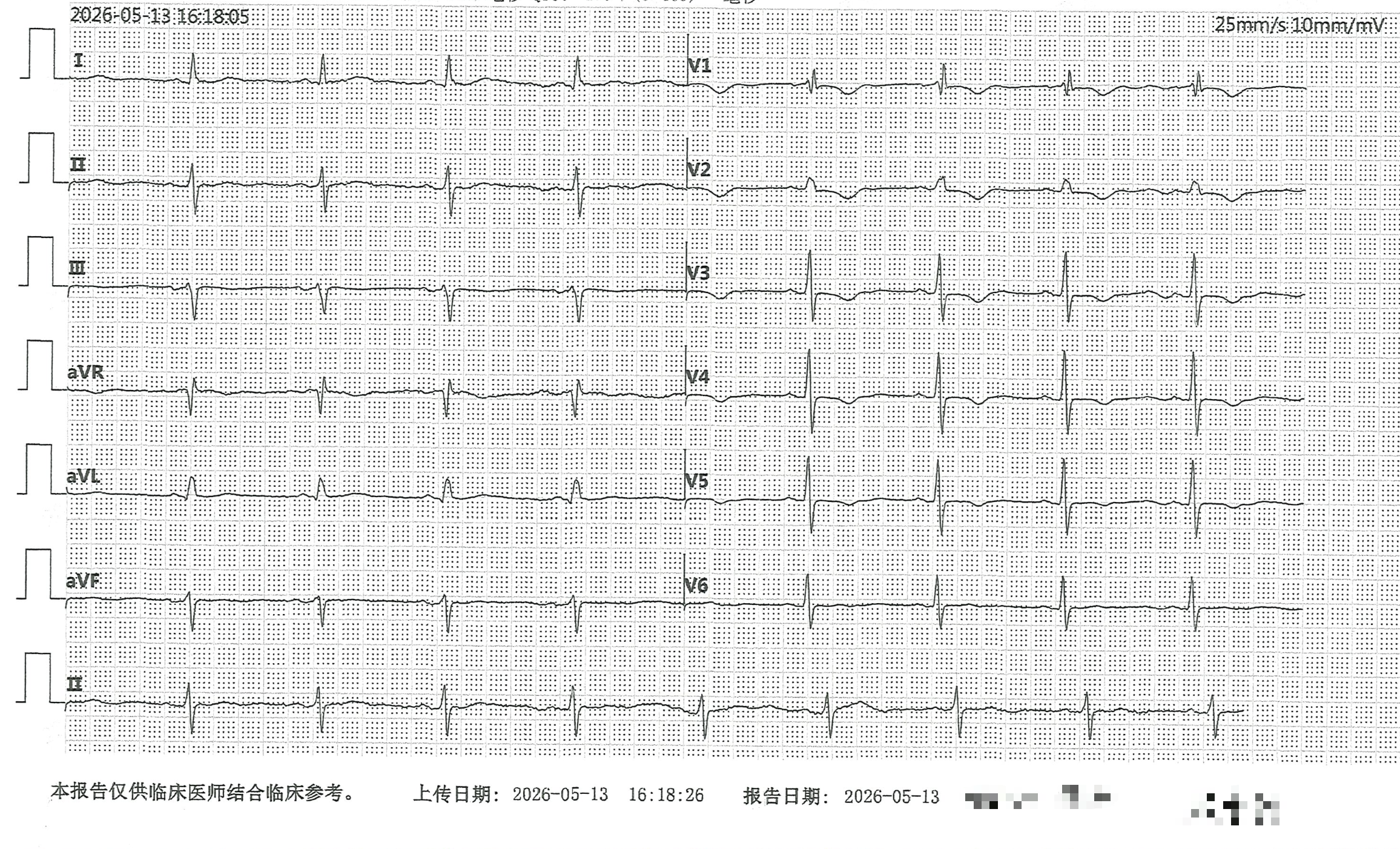}
    \vspace{3mm}
    \includegraphics[width=0.8\textwidth]{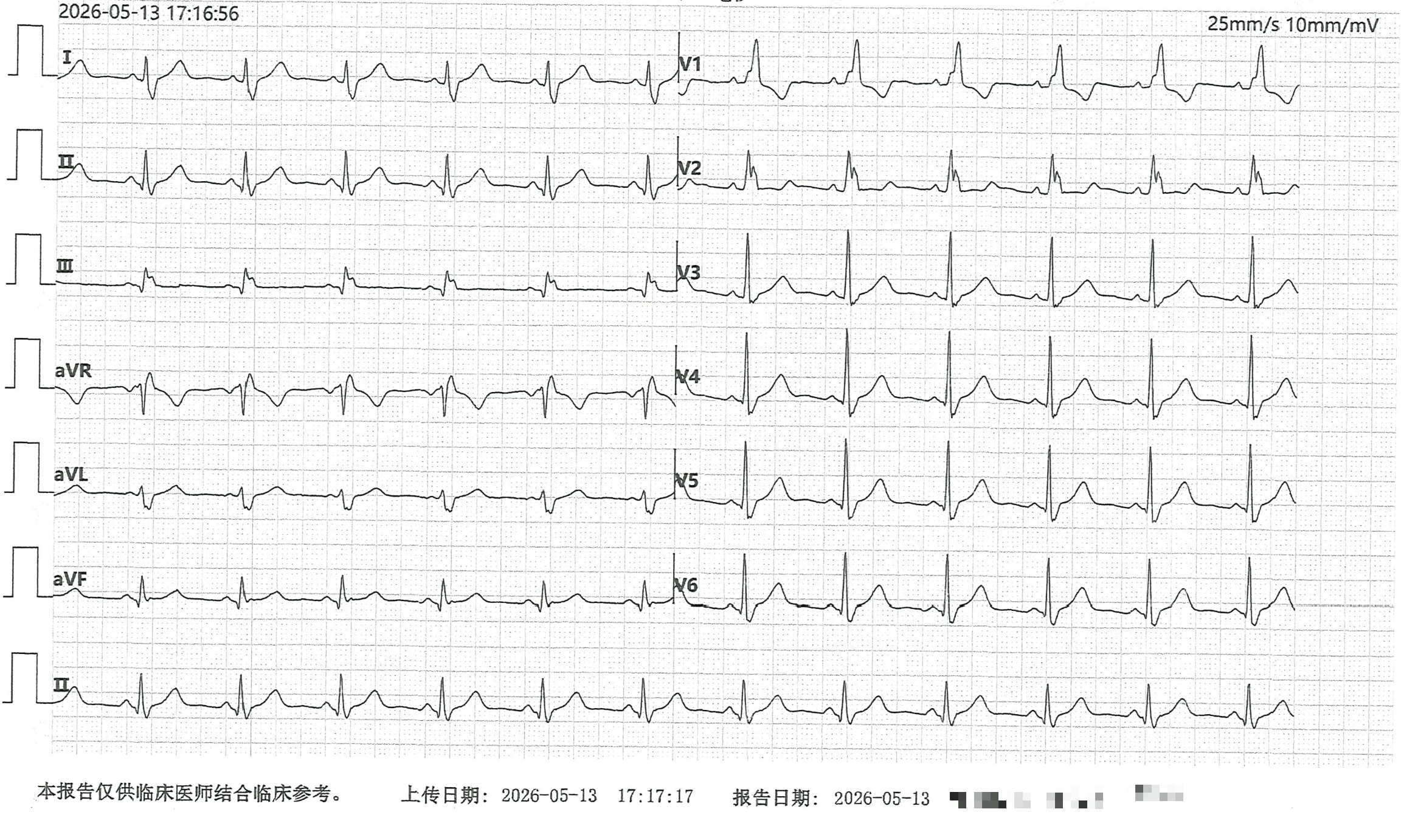}

    \caption{Examples of anonymized paper ECG images from the real-world cohort.}
    \label{fig:real-world-ecg-examples}
\end{figure*}

We collected a total of 312 paper ECGs from The First Affiliated Hospital with Nanjing Medical University between March 2026 and June 2026. 
The images were obtained using a Ricoh 8001 scanner. To ensure data de-identification and protect patient privacy, the top section of each scanned image was cropped to remove machine-generated ECG parameters and personally identifiable information (PII), including patient names, inpatient numbers, and ages. The processed images feature an asynchronous $6 \times 2$ layout, retaining the complete 12-lead ECG waveforms, and the acquisition timestamps.
Two  examples of these images are illustrated in Figure~\ref{fig:real-world-ecg-examples}.

The preliminary diagnostic opinions were provided by senior cardiologists. The image acquisition procedure was executed by a second-year graduate student, while the final data annotation and labelling were conducted by a Chief Physician.
Each original image was transformed to RGB and resized to fit within a $384 \times 224$ pixel canvas while preserving its original aspect ratio. 
We evaluated our model using a nested five-fold cross-validation strategy. For each outer iteration, one fold was held out for testing, while the remaining data underwent an another five-fold split to form the training and validation sets.

\section{Baselines and implementations}
The pretraining experiments comprised five baseline methods. SimCLR, MoCo, DINO, and BYOL were implemented as self-supervised learning baselines. 
All methods were pretrained using the generated PTB-XL ECG image dataset with an input resolution of $224 \times 224$. 
CvT-13 was adopted as the image encoder and initialized with ImageNet-pretrained weights. The random seed was fixed at 3407, and the training batch size was set to 64. All pretraining experiments were conducted on a single NVIDIA RTX PRO 6000 GPU with 96 GB of memory.
\begin{table*}[h]
\centering
\scriptsize
\setlength{\tabcolsep}{4pt}
\renewcommand{\arraystretch}{1.08}
\caption{Pretraining settings of baseline methods.}
\label{tab:baseline_pretraining_settings}
\resizebox{\textwidth}{!}{%
\begin{tabular}{lcccll}
\toprule
Method & Max epoch & Actual epoch & Pretraining LR & Optimizer & Early stopping strategy \\
\midrule
SimCLR & 50 & 38 & $3\times10^{-2}$ & SGD & Patience = 5; validation contrastive loss \\
MoCo & 80 & 26 & $3\times10^{-2}$ & SGD & Patience = 5; validation contrastive loss \\
DINO & 50 & 50 & $5\times10^{-4}$ & AdamW & Patience = 5; validation self-distillation loss \\
BYOL & 50 & 16 & $3\times10^{-2}$ & SGD & Patience = 8; validation BYOL loss \\
\bottomrule
\end{tabular}%
}
\end{table*}

For downstream fine-tuning, in addition to the five pretrained baseline methods, we further included two image models (PULSE and GEM), each containing 303.5M parameters. All methods were fine-tuned on the same downstream datasets with 1\% and 10\% training data, using five random seeds from 3407 to 3411.

\section{Distances for ordinal pairs}
\label{app:image-distances}

\textbf{Geometric distance.} 
Given an image $\mathbf{x}$, we extract its binary Canny edge map, where $\mathcal{P}$ denotes the set of edge-pixel coordinates. The geometric discrepancy between images $\mathbf{x}_{a}$ and $\mathbf{x}_{b}$ is computed via the symmetric Chamfer distance:
\begin{equation}
d_g(\mathbf{x}_{a},\mathbf{x}_{b}) = \frac{1}{2D} \left( \frac{1}{|\mathcal{P}_{a}|}\sum_{\mathbf{u}\in\mathcal{P}_{a}}\min_{\mathbf{v}\in\mathcal{P}_{b}}\|\mathbf{u}-\mathbf{v}\|_2 + \frac{1}{|\mathcal{P}_{b}|}\sum_{\mathbf{v}\in\mathcal{P}_{b}}\min_{\mathbf{u}\in\mathcal{P}_{a}}\|\mathbf{v}-\mathbf{u}\|_2 \right),
\label{eq:canny-chamfer-distance}
\end{equation}
where $D$ is a distance normalization factor.

\textbf{CIELAB distance.} 
To evaluate variations in paper color, illumination, and contrast, we map images into the CIELAB color space. For each channel $c\in\{L,a,b\}$, we compute the one-dimensional Wasserstein-1 distance, denoted as $W_{1,c}$, between their marginal cumulative histograms. The final appearance distance is the average across the three channels:
\begin{equation}
d_c(\mathbf{x}_{a},\mathbf{x}_{b}) = \frac{1}{3} \sum_{c\in\{L,a,b\}} W_{1,c}(\mathbf{x}_{a},\mathbf{x}_{b}).
\label{eq:lab-wasserstein-distance}
\end{equation}
This serves as a computationally efficient metric rather than an exact joint three-dimensional optimal-transport distance.


\section{Ablation study}

We ablate the rank-loss weight \(\lambda\) while keeping all other pretraining and evaluation settings fixed. Specifically, all experiments use \texttt{set} positive aggregation, a set-level multi-positive InfoNCE temperature of 0.10, a ranking margin of \(m=0.10\), an ordinal-edge separation parameter of \(\eta=0.05\), an anchor batch size of 64, and seed 3407. The encoder is optimized using an initial learning rate of 0.03, a minimum learning rate of \(10^{-5}\), one warm-up epoch, momentum of 0.9, and weight decay of \(10^{-4}\). Online augmentation is disabled. For evaluation, the encoder is frozen and a single linear head is trained with a learning rate of 0.1.
Table~\ref{tab:rank-weight-stagewise-ablation} shows that a moderate rank-loss weight of \(\lambda=0.5\) provides the best overall balance between discriminative performance and robustness to progressive degradation.

\begin{table*}[h]
    \centering
    \scriptsize
    \setlength{\tabcolsep}{1.8pt}
    \caption{Ablation study.}
    \label{tab:rank-weight-stagewise-ablation}
    \resizebox{\textwidth}{!}{%
        \begin{tabular}{l*{18}{c}}
            \toprule
            \multirow{2}{*}{Pareto weight}
            & \multicolumn{3}{c}{Clean Paper ECG}
            & \multicolumn{3}{c}{Severity 1}
            & \multicolumn{3}{c}{Severity 2}
            & \multicolumn{3}{c}{Severity 3}
            & \multicolumn{3}{c}{Severity 4}
            & \multicolumn{3}{c}{Avg.} \\
            \cmidrule(lr){2-4}
            \cmidrule(lr){5-7}
            \cmidrule(lr){8-10}
            \cmidrule(lr){11-13}
            \cmidrule(lr){14-16}
            \cmidrule(lr){17-19}
            & AUROC & AUPRC & F1
            & AUROC & AUPRC & F1
            & AUROC & AUPRC & F1
            & AUROC & AUPRC & F1
            & AUROC & AUPRC & F1
            & AUROC & AUPRC & F1 \\
            \midrule
            $0$
            & 0.835 & 0.637 & 0.611
            & 0.836 & 0.638 & 0.611
            & 0.831 & 0.631 & 0.607
            & 0.820 & 0.618 & 0.597
            & 0.803 & 0.592 & 0.576
            & 0.825 & 0.623 & 0.600 \\

            $\mathbf{0.5}$ (RobECG-CL)
            & \textbf{0.841} & \textbf{0.649} & 0.619
            & 0.839 & \textbf{0.649} & \textbf{0.620}
            & \textbf{0.837} & \textbf{0.644} & \textbf{0.617}
            & \textbf{0.823} & \textbf{0.624} & \textbf{0.607}
            & \textbf{0.805} & \textbf{0.599} & 0.575
            & \textbf{0.829} & \textbf{0.633} & \textbf{0.607} \\

            $1.0$
            & 0.838 & 0.644 & 0.617
            & 0.836 & 0.641 & 0.614
            & 0.834 & 0.638 & 0.614
            & \textbf{0.823} & 0.621 & 0.599
            & 0.802 & 0.590 & \textbf{0.581}
            & 0.827 & 0.627 & 0.605 \\

            $2.0$
            & 0.840 & 0.648 & \textbf{0.620}
            & \textbf{0.840} & 0.646 & 0.617
            & 0.836 & 0.640 & 0.615
            & 0.820 & 0.617 & 0.594
            & 0.787 & 0.570 & 0.551
            & 0.825 & 0.624 & 0.599 \\
            \bottomrule
        \end{tabular}%
    }
\end{table*}

\section{Supplementary related work}

Large pretrained models have improved transfer learning for waveform-based ECG analysis. ECG-FM combines masked reconstruction and contrastive learning to learn label-efficient representations \cite{mckeen2024ecg}. ECGFounder uses more than ten million cardiologist-annotated ECG recordings to support transfer across clinical tasks and data sources \cite{li2024electrocardiogram}. CSFM further uses masked generative pretraining on heterogeneous cardiac signals and associated text, enabling transfer across tasks, lead configurations, and sensing devices \cite{gu2026cardiac}. These models provide strong references for low-label transfer, operating on digital waveforms.

Image-based ECG models instead learn directly from ECG images. ECG-DualNet performs multilabel prediction from digitally rendered ECG images, while PRESENT-SHD uses ECG images generated with many predefined formats for structural heart disease screening \cite{sangha2022automated,dhingra2025present}. The PhysioNet Challenge 2024 extended ECG image analysis to synthetic and real images with scanning and imaging artifacts \cite{reyna2024challenge,reyna2024ecgimage}. 
In parallel, visual self-supervised learning provides a general approach to transferable representation learning: SimCLR and MoCo learn augmentation-invariant features using contrastive objectives, and BYOL and DINO learn from paired views without explicit negative samples \cite{chen2020simclr,he2020moco,grill2020byol,caron2021dino}. RobECG-CL learns consistent representations for progressively degraded views of the same ECG recording while preserving their relative degradation relationships.

\end{document}